\documentclass[lettersize,journal]{IEEEtran}

\usepackage{amsmath,amsfonts}
\usepackage{algorithmic}
\usepackage{algorithm}
\usepackage{array}
\usepackage[caption=false,font=normalsize,labelfont=sf,textfont=sf]{subfig}
\usepackage{textcomp}
\usepackage{stfloats}
\usepackage{url}
\usepackage{verbatim}
\usepackage{graphicx}
\usepackage{cite}
\usepackage{times}
\usepackage{colortbl}
\usepackage{epsfig}
\usepackage{graphicx}
\usepackage{amsmath}
\usepackage{amssymb}
\usepackage{enumitem} 
\usepackage[linesnumbered,ruled,vlined,algo2e]{algorithm2e}
\usepackage{multirow}
\usepackage{threeparttable}
\usepackage{stfloats}
\usepackage{booktabs}
\usepackage{arydshln}
\usepackage{bm}
\usepackage{bbding}
\usepackage[table,dvipsnames]{xcolor} 

\usepackage{ulem}
\usepackage[colorlinks]{hyperref}
\colorlet{colorFst}{Green!15}       
\colorlet{colorSnd}{SpringGreen!35} 
\colorlet{colorTrd}{Yellow!15}      
\colorlet{colorLow}{darkgray!50}    
\newcommand{\fs}{\cellcolor{colorFst}}   
\newcommand{\nd}{\cellcolor{colorSnd}}      
\newcommand{\rd}{\cellcolor{colorTrd}}      
\newcommand{\rfs}{\rowcolor{colorFst}}   

\newcommand{\revised}{\color{black}}

\def\eg{\emph{e.g.}\xspace} 
\def\ie{\emph{i.e.}\xspace}

\begin{document}

\title{Searching from Area to Point: A Semantic Guided Framework  with Geometric Consistency for Accurate Feature Matching}

\author{Yesheng Zhang~\IEEEmembership{Student Member,~IEEE}, Xu Zhao,~\IEEEmembership{Member,~IEEE}
\thanks{Yesheng Zhang and Xu Zhao are with the Department of Automation, School of Electronic Information and Electrical Engineering, Shanghai Jiao Tong University, Shanghai 200240, China.(email: preacher@sjtu.edu.cn; zhaoxu@sjtu.edu.cn, \textit{Corresponding author: Xu Zhao}).}

}

\markboth{Journal of \LaTeX\ Class Files,~Vol.~14, No.~8, August~2021}%
{Shell \MakeLowercase{\textit{et al.}}: A Sample Article Using IEEEtran.cls for IEEE Journals}

\IEEEpubid{0000--0000/00\$00.00~\copyright~2021 IEEE}

\maketitle

\begin{abstract}
Feature matching plays a pivotal role in computer vision applications. To achieve efficient and accurate matching, current methods commonly employ a \textit{coarse-to-fine} strategy, which establishes an intermediate search space preceding point matches. However, the difficulty in establishing dependable intermediate search spaces poses a limitation on the overall matching performance of existing feature matching methods. To address this issue, this paper proposes the integration of robust semantic priors in the intermediate search space and introduces a semantic-friendly search space called semantic area matches for precise feature matching.
The semantic area matches comprise matched image areas with significant semantic content, which can be robustly attained due to the semantic invariance against matching noise. Moreover, it facilitates point matching by reducing the redundancy and enables high-resolution input.
To adopt this search space, we introduce a hierarchical feature matching framework called \textit{Area to Point Matching} (\textbf{A2PM}), which involves identifying semantic area matches between images and subsequently conducting point matching on these area matches.
Furthermore, we present the \textit{Semantic and Geometry Area Matching} (\textbf{SGAM}) method to implement this framework, which leverages semantic priors and geometric consistency to establish precise area and point matches between images.
Through the adoption of the A2PM framework, SGAM demonstrates substantial and consistent enhancements in the performance of sparse, semi-dense, and dense point matchers in extensive point matching (up to $+29.16\%$) and pose estimation (up to $+13.01\%$) experiments. The code is publicly available at \href{https://github.com/Easonyesheng/SGAM}{https://github.com/Easonyesheng/SGAM}.
\end{abstract}


\section{Introduction}
\IEEEPARstart{F}{eature} matching is a fundamental task in computer vision, which serves as the basis of a wide range of vision applications, such as simultaneous localization and mapping \cite{SLAMSurvey}, structure from motion \cite{COLMAP} and image alignment \cite{glampoints}. 
Despite its status as a well-studied task, accurately determining the projections of a single 3D point in two different viewpoints continues to pose challenges.
These challenges arise from \textbf{matching noises}, such as potential extreme viewpoints, light variations, repetitive patterns, and motion blur, all of which result in the limited matching accuracy.

\begin{figure}[!t]
\centering
\includegraphics[width=\linewidth]{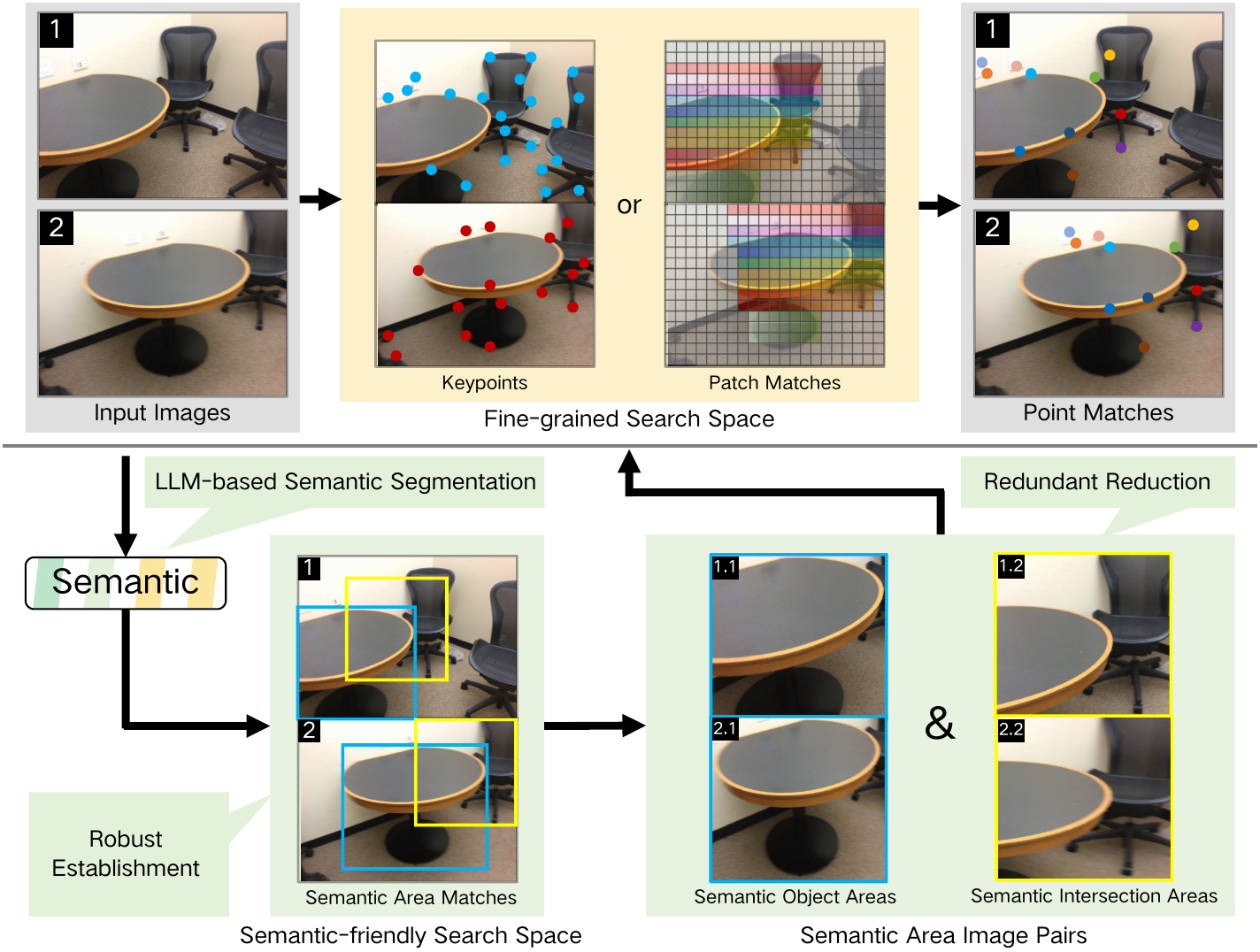}
\vspace{-1.2em}
\caption{\textbf{The proposed semantic-friendly search space of feature matching.} This search space, termed as \textbf{semantic area matches}, can be robustly established by leveraging semantic invariance, thereby reducing redundant computations in feature matching. As a result, fine-grained search spaces can be reliably established within the area image pairs, consequently enhancing matching accuracy.}
\vspace{-1.8em}
\label{fig_pre}
\end{figure}

Current feature matching methods are divided into sparse, semi-dense and dense methods~\cite{matchsurvey}.
Despite differences in specific techniques, narrowing the search space by means of hierarchical matching is the consensus of these methods. 
Typically, these methods begin by establishing intermediate search spaces for point matching between images, followed by the point matching within these spaces.
Specifically, in sparse methods~\cite{sift,alike}, the keypoint set is first detected in images, from which correspondences are subsequently achieved. While finding point matches from the keypoint set is easy in the sparse methods, detecting keypoints even with deep CNN \cite{superpoint, aslfeat}, however, suffers from inaccurate and failed detection caused by matching noises. 
\IEEEpubidadjcol
In semi-dense methods~\cite{loftr,ASpan}, sparse image patch matches are initially found by dense feature comparison, with point matches refined from these patches. Similarly, dense methods~\cite{NCN,dkm} progressively refine dense warps from coarse to fine feature maps, utilizing dense patch matches between images as the intermediate search space.
However, the patch matching relies on dense feature comparison, which leads to error-prone computation on irrelevant features and limited input resolution, consequently impacting the overall accuracy of semi-dense and dense methods. 

To address the redundant computation, two-stage matching methods~\cite{MKPC,OETR} propose to achieve the co-visible area (or \textit{region}) match between images as an intermediate search space of feature matching. Nonetheless, establishing this search space still heavily relies on feature comparison and suffers from matching noises.
On the other hand, PATS~\cite{pats} proposes a multi-stage patch matching by iteratively segmenting images into smaller patches and adjusting their scales according to match results. While PATS mitigates limited input resolution, it still involves unnecessary feature comparison, which restricts matching accuracy.

Following the coarse-to-fine matching idea, we emphasize the need to establish an improved intermediate search space to address the accuracy challenges in current feature matching methods. When designing a matching search space, two essential aspects warrant consideration: the ease of constructing the search space reliably and the precision of following matching within it. We propose the incorporation of semantics into the search space design to effectively tackle these dual concerns.
Previous work~\cite{sfd2,topicfm} have introduced semantic into feature matching to leverage the semantic invariance against matching noises. Nonetheless, they maintain the original search space and utilize semantic to enhance patch or keypoint features, leading to conflicting fine-grained search space and semantic labeling. In other words, current semantic perception methods struggle to accurately define boundaries between different semantics~\cite{SEEM}. However, these boundaries are precisely where matching search spaces tend to cluster, due to significant changes in image features. Thus, achieving fine-grained search spaces enhanced by semantics can be prone to semantic errors. 
Conversely, we suggest a coarse yet robust search space, termed as semantic area matches (Fig.~\ref{fig_pre}), to enable better integration with semantics. 

This area-level search space determines its size according to internal semantic cues, ensuring the inclusion of an adequate amount of semantic information within a condensed area to differentiate it from the entire image. Specifically, we propose two typical semantic areas: one encompassing an entire object and the other representing the intersection of multiple semantic entities. Thus, benefiting from semantic robustness in matching, the search space can be established without being influenced by semantic errors at the boundaries.
Meanwhile, irrelevant feature computation is circumvented in the matched semantic areas, thus allowing for high resolution input and improving the accuracy of following matching phases.
To actualize this search space, we introduce Semantic Area Matching (SAM) to detect and match semantic areas across images, powered by the advanced Large Language Model-based (LLM) semantic segmentation method~\cite{SEEM}.
With precise semantic segmentation of images, we show that semantic area matches can be easily achieved by hand-crafted semantic features, leading to a notable enhancement in matching accuracy.

\begin{figure*}[!t]
\centering
\includegraphics[width=\linewidth]{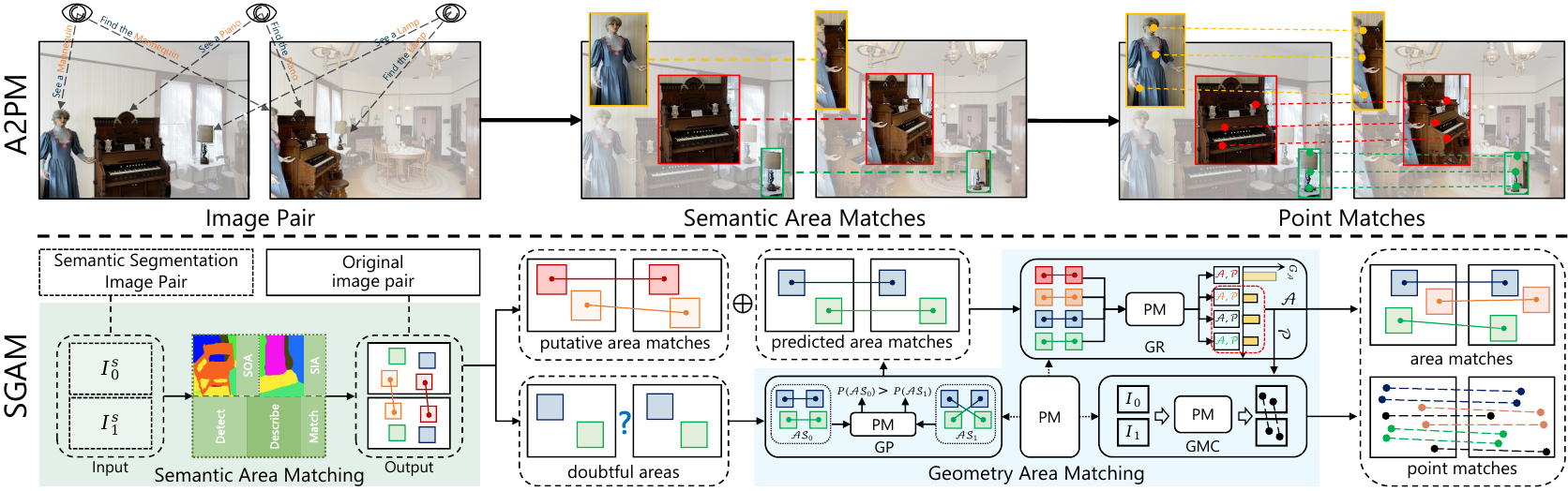}
\caption{\textbf{Overview of the proposed feature matching method.} \textbf{(i) Top:} The proposed \textit{Area to Point Matching} (A2PM) framework initially identifies semantic area matches between images and then conducts point matching within the matched areas. \textbf{(ii) Bottom:} {We propose \textit{Semantic and Geometry Area Matching} (SGAM) method, which encompasses {\textit{Semantic Area Matching} (SAM)} and {\textit{Geometry Area Matching} (GAM)}. The SAM leverages semantic segmentation to detect and match \textit{semantic object areas} (SOA) and \textit{semantic intersection areas} (SIA) between the images. Integrated with an off-the-shelf \textit{Point Matcher} (PM), the GAM comprises a \textit{Predictor} (GP) for determining true matches within doubtful areas, a \textit{Rejector} (GR) for filtering out false and poor area matches and a \textit{Global Match Collection} (GMC) module to further enhance the robustness under low semantic scenes, by collecting accurate global correspondences.}
}
\label{fig_A2PM}
\vspace{-1.2em}
\end{figure*}

Due to the image-level size of semantic area matches, they can serve as direct inputs for point matching, similar to overlap area match~\cite{OETR}. However, semantic area matches are more refined than the overlap area. It converts the original point matching task into multiple point matching tasks. We propose a matching framework, named Area to Point Matching (A2PM, Fig.~\ref{fig_A2PM} top), to formulate the matching process with semantic area matches. Initially, It establishes semantic area matches between images and then extracts these areas from the original images to perform point matching within them.
This framework offers several advantages: \textbf{1)} By re-cropping matched areas from high-resolution images, point matching benefits from more detailed inputs than the original. \textbf{2)} The decoupling of the search space establishment from point matching allows that the accuracy of various point matching methods can be improved by the same area matching method.

However, there is no free launch in feature matching when leveraging semantic. 
Although SAM effectively detects and matches most semantic areas between images, the inherent abstraction property in semantics overlooks local details. This can lead to \textit{semantic ambiguity} during matching, particularly when distinct instances coexist within the image. Therefore, SAM may identify doubtful areas that cannot be confidently matched.
Besides, the semantic ambiguity may also lead to erroneous area matches in SAM, adversely affecting feature matching.
To tackle this challenge, we turn to the intrinsic geometric properties of area matches.
In particular, considering that area matches across images signifies the same 3D entity, the constraint of epipolar geometry naturally applies.
Moreover, area matches in the same image pair must adhere to the same constraint, \ie \textit{geometric consistency}, which can be employed to address the semantic ambiguity.
Hence, we establish the geometric consistency of area matches by utilizing the epipolar geometry constraint of point matches within these areas.
It enables the proposed \textit{Geometry Area Matching} (GAM) to integrate a point matcher for accurate refinement of area matches. 
In practice, the GAM first predicts true area matches from doubtful candidates generated by SAM (GAM Predictor, GP).
Subsequently, all area matches undergo filtering by the GAM Rejector (GR) to identify superior matches with geometric consistency. 
The point matches within the accurate area matches are obtained at the same time.
Furthermore, to handle the correspondence aggregation issue in less-semantic scenes, the Global Match Collection module (GMC) is incorporated in GAM, which involves collecting additional point matches globally based on the geometry consistency of inside-area correspondences. The GMC ensures the generation of well-distributed matches, advantageous for downstream tasks.
Through the combination of SAM and GAM, our \textit{Semantic and Geometry Area Matching} (SGAM, Fig. \ref{fig_A2PM} bottom) is capable of achieving accurate area and point matching between images.

In sum, our contributions are as follows:
\begin{enumerate}[itemsep=2pt,topsep=0pt,parsep=0pt]
    \item Introduction of a semantic-friendly intermediate search space for feature matching, called semantic area matches, accompanied by a corresponding matching framework named A2PM. This framework involves the initial establishment of semantic area matches between images, and then performs point matching within these area image pairs, improving the matching accuracy ultimately.
    \item To implement the A2PM framework, we propose the SGAM approach, which consists of two components: SAM, responsible for identifying putative area matches according to semantics, and GAM, {which obtains precise area and point matches} by ensuring geometry consistency.
    \item Utilizing LLM-based semantic segmentation method, SGAM shows consistent improvement on matching accuracy for sparse, semi-dense and dense matching methods, thereby leading to impressive pose estimation performance on various indoor and outdoor datasets.
\end{enumerate}

\section{Related Work}

\subsection{Sparse Matching} 
{Sparse matching relies on detected keypoints and their descriptors~\cite{sift,orb}. 
Through the nearest neighbor search based on descriptor distances, point matches can be established between images.}
In the age of deep learning, recent work~\cite{aslfeat,superpoint,r2d2,disk} utilize deep CNN to achieve better learning feature.
Specifically, SuperPoint~\cite{superpoint} is early in providing feature detection and description networks and outperforms conventional methods.
{Subsequent work \cite{d2net,aslfeat,alike} leverage a unified network to detect and describe feature.}
At the same time, detached learning detection \cite{keyNet,rotation,eqkd} and description \cite{ng2022ninjadesc,pump,decoupling} are proposed as well. 
After feature detection, point match searching and outlier rejection are also advanced by recent learning methods \cite{superglue,T-Net,OANet,OR_TIP,Msa,PGF}.
{
Essentially, this framework relies on extremely fine-grained search space establishment, i.e. the feature points, to achieve accurate point matching.
However, feature point detection poses significant challenges in scenes with low texture, repetitive patterns, extreme changes in illumination and scale, which ultimately leads to a decline in performance.}
In contrast, our method appropriately reduces the search space to semantic area matches, which are robust due to their semantic invariance.

\subsection{Semi-dense and Dense Matching}
In order to avoid detection failure, semi-dense and dense framework is proposed \cite{NCN,loftr,cotr}, aiming at jointly trainable feature detection, matching and outlier rejection to establish point matches directly from image pairs. 
{Initially, 4D CNN is used to extract and compare image features densely~\cite{NCN,sparseNCN,dual-resolution} in dense methods. Recent DKM~\cite{dkm} constructs a Gaussian Process using CNN and achieves leading performance. 
Owing to limited receptive field \cite{cotr} of CNN, COTR \cite{cotr} incorporates Transformer \cite{tf} to process dense feature extracted by CNN. To alleviate the high computational cost, semi-dense methods such as LoFTR~\cite{loftr} and its variants~\cite{QT,ASpan} suggest selecting sparse patch matches after dense computation to refine point matches.
Nonetheless, the dense feature comparison in these methods leads to redundant computation, thus limiting the resolution of input images. Furthermore, the redundant search space introduces noise from non-overlapping areas in the image pair, resulting in degraded overall precision. As a solution, we propose to utilize semantics to reduce redundant computation through our semantic area matches. 
Due to the higher resolution and less noise in the area matches than original images, the precision of subsequent inside-area point matching is improved.

\subsection{Coarse-to-Fine Matching}
Recent matching methods commonly adopt a coarse-to-fine matching approach, which is deemed as a consensus in the field. Within this hierarchical process, intermediate search spaces play a critical role in determining the overall matching accuracy, and these spaces vary across different methods.
Sparse methods typically identify a keypoint set as the search space for point matching, but struggle to achieve fine-grained keypoints.
In semi-dense methods~\cite{loftr,QT,ASpan}, patch-level matches are established through feature attention of the coarse level matching, which serve as the search space for the fine level matching. This approach significantly reduces time consumption compared to the dense counterpart \cite{cotr}.
On the other hand, dense methods~\cite{dkm,sparseNCN,dual-resolution} refine dense warps from coarse to fine feature maps, where dense correspondences between coarse feature maps can be viewed as dense patch matches between original images, making patch matches the intermediate search spaces for dense methods as well.
Despite these advancements, the patch-level search spaces lack a clear association with image context, thus requiring expensive feature comparison in establishment.
The accuracy issue in the coarse matching also persists due to low input resolution and error-prone redundant computations.
While PATS~\cite{pats} is proposed to extract more accurate features from equally cropped image patches with high resolution, the presence of redundant feature comparisons continues to limit matching accuracy.
Recent overlap estimation methods~\cite{OETR,MKPC} utilize stage-one matching on the entire images to achieve the overlap between images.
Then the stage-two point matching is performed inside the overlap area images.
However, individual overlap area may be too coarse for precise point matching and the overlap establishment in these methods are expensive.
Conversely, our method offers a semantic-aware search space, \ie the semantic area matches.
The incorporation of semantic enables robust establishment and efficient reduction of redundant computation of this search space, leading to impressive matching accuracy.

\section{Methodology}
In this section, we first formulate the A2PM framework and its geometry properties (Sec.~\ref{sec_form}).
Then, we propose the \textit{Semantic Area Matching} (Sec.~\ref{sec:SAM}) with hand-crafted detection and description to establish putative semantic area matches, utilizing semantic robustness to overcome matching noises.
To refine the area matches, we leverage geometry consistency formulated in Sec.~\ref{sec:geo-form} and propose the \textit{Geometry Area Matching} (Sec.~\ref{sec:GAM}). 
Finally, the implementation of A2PM framework by combining SGAM with any point matcher is illustrated (Sec.~\ref{sec:FI}). 
We provide a summarise of symbols used in Tab.~\ref{table_sym} of the appendix.

\subsection{Formulation}\label{sec_form}
The formulation section includes the detailed description about the the proposed A2PM framework and the proposed geometry consistency of area matching.

\subsubsection{A2PM Framework}
Generally, given an image pair ($I_0, I_1$), an area matching method $AM$ and a point matching method $PM$, the A2PM framework ($\mathcal{M}_A$) is responsible for connecting the area matching and point matching to achieve the final point matches accurately:
\begin{equation}
    \mathcal{P} = \mathcal{M}_A (I_0, I_1, AM, PM).
\end{equation}
The output $\mathcal{P} = \{q^m, p^m\}_m^M$ is the set of point matches ($q\in I_0, ~p \in I_1$ are correspondences).
Specifically, the area matching can be formulated as follows.
Suppose two matched areas in $I_0,I_1$ are respectively $\{\alpha_i\}_i$ and $\{\beta_j\}_j$. The area match is represented as $\mathcal{A}_{i,j}=(\alpha_i, \beta_{j})$.
In area matching, $N$ pairs of area matches can be achieved by the $AM$:
\begin{equation}
    \{\mathcal{A}_{i,\pi(i)}\}_{i}^{N} = AM(I_0, I_1),
\end{equation}
where $\pi(i): \mathbb{R} \rightarrow \mathbb{R}$ is the index mapping between matched areas. Then, point matches inside each area match can be found to compose the final point matches:
\begin{equation}
    \mathcal{P} \leftarrow \{PM(\mathcal{A}_{i,\pi(i)})\}_{i}^{N}.
\end{equation}
Due to the higher resolution and less redundancy of semantic area matches than the original input images of $PM$, the accuracy of final point matches is improved.

\subsubsection{Geometry Consistency of Area Matching}\label{sec:geo-form}
In order to leverage the intrinsic geometry property to improve matching accuracy and robustness, we proceed to formulate the geometry consistency of area matching.
Since point-level geometry constraint is formed completely~\cite{MVG_book}, we utilize the epipolar geometry of point matches within areas to construct the area-level geometry consistency.
First, the correspondences {$\mathcal{P}_i = \{(q^m_i,p_i^m)\}_m^M$} in $\mathcal{A}_{i,\pi(i)}$ can be achieved by $PM$ and the fundamental matrix $F_i$ can be calculated as well.
Then we form the geometry consistency between $\mathcal{P}_i$ and $F_i$ by Sampson distance \cite{MVG_book}:

\begin{equation}\label{eq_selfGC}
\begin{aligned}
    d_{i,i}& = \sum_{m}^M \frac{(p_i^{mT} F_i q_i^m)^2}{(F_i q_i^m)_1^2 + (F_i q_i^m)_2^2 + (F_i^T p_i^m)_1^2 + (F_i^T p_i^m)^2_2} \\
    & = {\sum_{m}^M \hat{d}^m_{i, i}} = D(F_i, \mathcal{P}_i),
\end{aligned}
\end{equation}
where $(F_i q_i^m)_k$ represents the k-th entry of the vector $F_i q_i^m$. It should be ideally close to 0 and reflects the matching precision of $\mathcal{A}_{i,\pi(i)}$, as only the correct area match produces accurate point matches.
Similarly, we can infer the geometry consistency across area matches.
For two correct area matches $\{\mathcal{A}_{i,\pi(i)}, A_{j, \pi(j)}\}$ between the images, they should ideally yield the same fundamental matrix.
Thus the \textit{cross Sampson distance} ($d_{i,j}$) should be close to 0:
\begin{equation}\label{eq:d_ij}
\begin{aligned}
    d_{i,j} & = D(F_i, \mathcal{P}_j) \rightarrow 0.
\end{aligned}
\end{equation}
Therefore, within an area match set $\{\mathcal{A}_{i,\pi(i)}\}_{i}^{N}$, assuming the most of area matches are correct, the geometry consistency of a specific area match $\mathcal{A}_{i,\pi(i)}$ can be formulated as:
\begin{equation} \label{eq_crossGC}
    G_{\mathcal{A}_{i,\pi(i)}} =\frac{1}{N} \sum_{j}^{N} d_{i,j}
\end{equation}
Thus, the $G_{\mathcal{A}_{i,\pi(i)}}$ can reflect the matching accuracy of $\mathcal{A}_{i,\pi(i)}$ and the smaller the higher area matching precision.

\subsection{Semantic Area Matching}\label{sec:SAM}

\begin{figure}[!t]
\centering
\includegraphics[width=\linewidth]{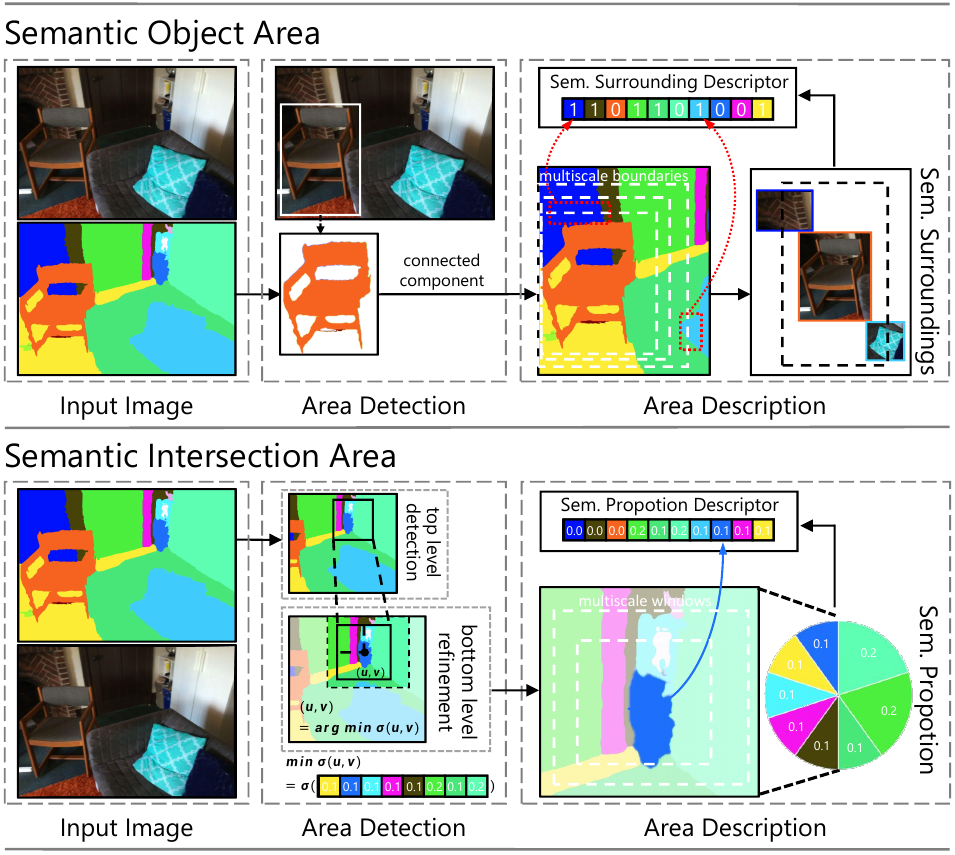}
\caption{\textbf{{Semantic Area Matching} (SAM).} Two types of semantic areas are proposed by SAM. Both of them are detected and described by hand-crafted semantic feature. Then area matches are established by nearest neighbour search based on descriptor distance.}
\vspace{-1em}
\label{fig_SAM}
\end{figure}

To find semantic area matches between images, we propose Semantic Area Matching (SAM), which adopts a \textit{detection-and-description} manner similar to sparse point matching~\cite{superpoint}.
Particularly, we first propose two typical semantic areas with the goal of achieving a better integration between semantics and search space.
The first area is an object-centric area, termed as \textit{Semantic Object Area} (SOA), where the textured surface and prominent edges of the inside object favour point matching. 
However, some objects (\eg objects very close to the camera) are so large in the image that the sizes or aspect ratios of the corresponding areas are extremely large, leading to improper search space. Thus, we further propose the \textit{Semantic Intersection Area} (SIA), which consists of intersecting parts of multiple objects rather than an entire object, to efficiently grab solid features in the intersection of above large objects. 
Afterwards, we illustrate the \textit{detection-and-description} matching processes of SOA and SIA (Fig.~\ref{fig_SAM}). Given semantic segmentation of images ($I^s_i$), both of them leverage hand-crafted semantic features to achieve area matches with sufficient accuracy, benefiting from semantic robustness in matching.

\subsubsection{Matching of SOA}\label{sec:soam} 
Detecting semantic object areas can be accomplished by identifying the connected components with the object semantic in $I^s_i$ and utilizing their bounding boxes as the boundaries for the areas.
For achieving a sparse extraction, we merge the spatially proximate areas that exhibit the same semantic.

As SOA already contains the distinguishable object semantic information, its descriptor is designed to differentiate it from other instances of the same semantics in the image.
Since close instances are merged in detection, it is crucial to focus on distinguishing the spatially scattered instances, which are likely to have different surroundings.
Therefore, we propose the \textit{semantic surrounding descriptor} to differentiate instances by leveraging semantic information about their surroundings.
Similar to BRIEF~\cite{brief} descriptor in point matching, this descriptor is also a binary vector where each bit corresponds to a semantic present in the image pair. A bit with a value of $1$ indicates the presence of the semantic along the area sides, while a value of 0 indicates its absence, thereby representing the semantic context surrounding the area.
To enhance scale robustness, we propose the multiscale boundaries capture. By scaling the area boundary to varying degrees, more semantics can be captured (see Fig.~\ref{fig_SAM} top `Area Description' part.).

The SOAs are first matched directly by their inside object semantics. 
Subsequently, Hamming distances~\cite{brief} between the descriptors of initial matches are calculated.
Next, area matches sharing the same semantic are determined through the nearest neighbor search.
During this process, matches with descriptor distances larger than the threshold $T_{H}$ are rejected.
Some SOAs may be labeled as doubtful, due to similar descriptor distances (identified by threshold $T_{da}$) of multiple match candidates.
These doubtful areas will be handled in the geometric area matching.

\begin{algorithm2e}[t]\label{alg_gmr}
    \normalem
    \caption{Geometric Area Matching Rejector}
    \KwIn{$\mathcal{A}\mathcal{S}=\{\mathcal{A}_{i,\pi(i)}\}_{i}^{S}$}
    \KwOut{$\{\mathcal{A}_{i^*,\pi(i^*)}, \mathcal{P}_{i^*}\}_{i^*}^{T},{i^* \in [0, S)}, T\leq S$}

    \For{$\mathcal{A}_{i,\pi(i)}$ in $\mathcal{A}\mathcal{S}$}
    {
        perform point matching inside the area match: $\mathcal{P}_i=PM(\mathcal{A}_{i,\pi(i)})$\;
        calculate the fundamental matrix: $F_i$\;
        get the self-geometry consistency by Eq. \ref{eq_selfGC}: $d_{i,i}$\;
    }
    calculate the geometry consistency threshold: $T_{GR} = \phi \times \frac{1}{S} \sum_{i}^{S} d_{i,i} $\;
    \For{$\mathcal{A}_{i,\pi(i)}$ in $\mathcal{A}\mathcal{S}$}
    {
        calculate the $G_{\mathcal{A}_{i,\pi(i)}}$ by Eq.~\ref{eq_crossGC}\;
        \lIf{$G_{\mathcal{A}_{i,\pi(i)}} > T_{GR}$}{reject $\mathcal{A}_{i,\pi(i)}$}
    }
    Output the left area matches and their correspondences: $ \{\mathcal{A}_{i^*,\pi(i^*)}, \mathcal{P}_{i^*}\}_{i^*}^{T}, {i^* \in [0, S)}, T\leq S$\;
\end{algorithm2e}

\subsubsection{Matching of SIA}\label{sec:siam} 
The detection of SIA involves sliding a window across the $I_i^s$ to identify areas with abundant semantic specificity. 
In particular, the window size is set to the desired area size, and the slide step is half the window size. 
During the window sliding, areas with more than $3$ different semantics are collected as SIAs. 
Considering the expensive time cost of window sliding in original $I_i^s$, we employ a two-layer semantic pyramid, similar to classical image pyramid.
The top layer involves reducing the $I^s_i$ and window to scale $r$, performing initial detection.
The bottom layer is the original $I_i^s$ , used to further refine the area location. 
In the refinement, we first calculate the proportion of every semantics within each area. 
Then, we adjust the center of each area within a certain range (the area size) to ensure uniform proportion of different semantics in the area, by minimizing the variance of semantic proportion ($\sigma(u,v)$) within it (see Fig.~\ref{fig_SAM} bottom `Area Detection' part.).

The inside-area semantics are crucial to match SIAs. Therefore, we propose utilizing the semantic proportion, calculated during the detection process, as the SIA descriptor. Similar to SOA, this descriptor is also a vector, but it is not binary. Each element of the vector represents a specific semantic, with its value indicating the proportion of that semantic within the area.
To enhance the scale robustness, the descriptors are constructed on multiscale windows with the same center and then merged by taking their average.

Afterwards, SIA matches can be found by nearest neighbour search based on $l_2$ distance between descriptors. Similar to SOA matching, doubtful areas are identified by $T_{da}$ and inferior matches are rejected by $T_{l}$.

\subsection{Geometry Area Matching}\label{sec:GAM}
Although the SAM is effective in most cases, it tends to overlook local details in images, potentially leading to \textit{semantic ambiguity} when multiple instances are present in the image pair.
Especially when semantic surroundings of instances are also similar, SAM may obtain doubtful areas and incorrect area matches.
Fortunately, area matches, similar to point matches, are inherently constrained by epipolar geometry, which can be utilized to resolve the semantic ambiguity.
Hence, based on the formulated geometry consistency in Sec.~\ref{sec_form}, we propose \textit{Geometry Area Matching} (GAM) to refine the results of SAM and fulfill the A2PM framework. {GAM incorporates a Predictor (GP, Sec.~\ref{sec:GP}) to identify true matches in doubtful areas, a Rejector (GR, Sec.~\ref{sec:GR}) to eliminate inferior area matches and a Global Match Collection module (GMC, Sec.~\ref{sec:GMC}) to achieve uniformly distributed matches.}

\subsubsection{Geometry Area Matching Predictor}\label{sec:GP}
The GP aims to determine the true matches among multiple matching possibilities of doubtful areas.
Given doubtful areas {\small $\{\alpha_i\}_{i}^{H}, \{\beta_j\}_{j}^{R}, R\leq H$} in $I_0,I_1$ which can not be confidently matched by SAM, and assume $R$ correct area matches exist:
\begin{equation}
    \mathcal{A}\mathcal{S}_l = \{\mathcal{A}_{i,\pi_l(i)}\}_{i}^{R} = \{(\alpha_i, \beta_{\pi_l(i)})\}_i^R,
\end{equation}
where $\mathcal{A}\mathcal{S}_l$ is a set of area matches, {\small$\pi_l(i) \in [0,R)$} {is an index mapping between matched areas with $l$ indicating different area matching possibilities.}
There are totally {\small$L={H!}/{(H-R)!}$} matching possibilities ({\small $l \in [0,L)$}), and only one true area match set ($\mathcal{A}\mathcal{S}_{l^*}$) exists with the best geometry consistency, where every area match is correctly matched.
Thus, we first form the geometry consistency of any $\mathcal{A}\mathcal{S}_l$ based on Eq.~\ref{eq_crossGC}:
\begin{equation}
    G_{\mathcal{A}\mathcal{S}_l} = \frac{1}{R}\sum_{i}^{R} G_{\mathcal{A}_{i,\pi_l(i)}}.
\end{equation}
Then the likelihood of $\mathcal{A}\mathcal{S}_l$ can be represented as:
{
\begin{equation}\label{eq:GP}
    P(\mathcal{A}\mathcal{S}_l) = \exp(-G_{\mathcal{A}\mathcal{S}_l}).
\end{equation}
}
Therefore, the true match set $\mathcal{A}\mathcal{S}_{l^*}$ can be achieved by likelihood maximization:
\begin{equation}
    {As}_{{l}^*} = \arg \max_{l}~ P(\mathcal{A}\mathcal{S}_l).
\end{equation}
This can be solved by considering the whole density of $\mathcal{A}\mathcal{S}_l$ and choose the one with the maximum $P(\mathcal{A}\mathcal{S}_l)$.

\subsubsection{Geometry Area Matching Rejector}\label{sec:GR}
Following the prediction, the GR leverages geometry consistency to identify and reject potential false matches, thereby enhancing the accuracy of the area matching.
Given an area match set $\mathcal{A}\mathcal{S} = \{\mathcal{A}_{i,\pi(i)}\}_{i}^{S}$ achieved by SAM and GP, the geometry consistency of each $\mathcal{A}_{i,\pi(i)}$ can be measured by $G_{\mathcal{A}_{i,\pi(i)}}$ (Eq.~\ref{eq_crossGC}). 
Then, matches with $G_{\mathcal{A}_{i,\pi(i)}}$ exceeding a specific threshold $T_{GR}$ can be discarded as inaccurate.
In practice, the $T_{GR}$ is based on the mean self-geometry consistency (Eq.~\ref{eq_selfGC}) with a weight $\phi$. 
The point matcher is embedded in GR to acquire point matches. The specific process is illustrated in Algorithm~\ref{alg_gmr}.

\begin{algorithm2e}[t]\label{alg_gmc}
    \normalem
    \caption{Global Match Collection}
    \KwIn{$\{\mathcal{P}_i, \mathcal{A}_{i, \pi(i)}\}_{i}^T$, $I_0$, $I_1$, $T_{SP}$}
    \KwOut{${\mathcal{P}}_{g}$}

    calculate the Size Proportion of area matches in images: $SP_{\{\mathcal{A}_{i, \pi(i)}\}_{i}^{T}}$

    initialize the ${\mathcal{P}}_{g} = \varnothing$

    \uIf{$SP_{\{\mathcal{A}_{i, \pi(i)}\}_{i}^{T}} < T_{SP}$ }{
     calculate the fundamental matrix $F_{a}$ and mean \textit{Sampson distance} $\frac{1}{T}\sum_i^T d_{a,i}$ (by Eq.~\ref{eq:d_ij}) of \textit{inside-area} point matches $\{\mathcal{P}_i\}_{i}^T$\;
     achieve the global matches: $\mathcal{P}_g = PM(I_0, I_1)$\;
    \For{$(p^m_g,q^m_g)$ in $\mathcal{P}_g$}
    {
    get the single match \textit{Sampson distance} {${d}^m_{a,g}$\;}
    \lIf{${d}^m_{a,g} <= \frac{1}{T}\sum_i^T d_{a,i}$}{collect the match $(p^m_g,q^m_g)$ into ${\mathcal{P}}_{g}$}
        
    }
    }
    Output the collected global matches: ${\mathcal{P}}_{g}$\;
\end{algorithm2e}

\subsubsection{Global Match Collection}\label{sec:GMC}
The precision of point matches within area matches is guaranteed, termed as $\{\mathcal{P}_i, \mathcal{A}_{i, \pi(i)}\}_{i}^T$, as a result of the improved search spaces achieved through both semantic prior and geometric consistency.
However, the distribution of these matches depends on the specific scenes.
If less semantic information is available in the scene, there will be a small number of area matches.
Consequently, the point matches will cluster, which has a negative impact on the downstream tasks~\cite{MVG_book}.
To enhance the robustness against scenes with limited semantic information, we propose the Global Match Collection (GMC) module to collect global matches ($\mathcal{P}_{g}$) utilizing the geometry constraint of accurate inside-area point matches in scenes with less semantic content. These scenes are identified based on a size proportion threshold $T_{SP}$, which is the proportion threshold of the image occupied by matched areas in the image pair.
The detailed algorithm is presented in Algorithm \ref{alg_gmc}.

\subsection{Framework Implementation}\label{sec:FI}
The overall A2PM framework follows the steps outlined below.
Firstly, given the semantic segmentation of the input image pair ($I^{s}_0, I^s_1$), the framework obtains putative area matches ({\small$\{ \mathcal{A}^*_{i, \pi(i)}\}_{i}^{K}$}) and doubtful areas ({\small$\{\alpha_i\}_{i}^{H}$, $\{\beta_i\}_{i}^{R}$, $R\leq H$)} between the images using the SAM algorithm:
\begin{equation}
    \small
    \{ \mathcal{A}^*_{i, \pi(i)}\}_{i}^{K}, \{\alpha_i\}_{i}^{H}, \{\beta_i\}_{i}^{R} = SAM(I^{s}_0, I^{s}_1).
\end{equation}
Next, the doubtful areas are cropped from original images ($I_0, I_1$) and matched by a Point Matcher ($PM$) integrated in $GP_{PM}$ to achieve inside-area correspondences for geometry consistency calculation. Then, the true area matches ({\small$\{\mathcal{A}^*_{i, \pi(i)}\}_{i}^{R}$}) can be identified by $GP_{PM}$:
\begin{equation}
    \{\mathcal{A}^*_{i, \pi(i)}\}_{i}^{R} = GP_{PM}(\{\alpha_i\}_{i}^{H}, \{\beta_i\}_{i}^{R}, I_0, I_1).
\end{equation}
Afterwards, the accurate area and point matches inside them ({\small$\{\mathcal{A}_{i, \pi(i)}, \mathcal{P}_i\}_{i}^{T}$}) are achieved by $GR_{PM}$ integrated with $PM$ (The areas are cropped and subsequently matched by PM.):
\begin{equation}
    \{\mathcal{A}_{i, \pi(i)}, \mathcal{P}_i\}_{i}^{T} = GR_{PM}(\{A^*_{i, \pi(i)}\}_{i}^{K+R}, I_0, I_1).
\end{equation}
{In case of less-semantic scenes, more accurate point matches from full-image point matching using $PM$ are obtained by our $GMC_{PM}$ module:}
\begin{equation}
    \mathcal{P}_{g}^C = GMC_{PM}(\{\mathcal{A}_{i, \pi(i)}, \mathcal{P}_i\}_{i}^{T}, I_0,I_1, T_{SP}).
\end{equation}

Finally, the output point matches are merged by inside-area matches \{$\mathcal{P}_i\}_{i}^{T}$ and global matches $\mathcal{P}_{g}^C$,
which possess both high matching accuracy and uniform spatial distribution. It is noteworthy the $PM$ we adopted can be any point matching method. Therefore, our SGAM is able to universally improve the accuracy of sparse, semi-dense and dense point matchers, as shown in our experiments.

\section{Results}

\begin{table*}[t]
\centering
\caption{\textbf{Value results (\%) of MMA.} We report MMA with three thresholds under various matching difficulties. 
\textbf{Our SGAM} is applied on four baselines. {To show the impact of semantic accuracy to our method, we take three different semantic inputs: \colorbox{colorFst}{SGAM using ground truth (GT)}, \colorbox{colorSnd}{SGAM using SEEM-L} and \colorbox{colorTrd}{SGAM using SEEM-T}.}
The improvement achieved by SGAM is also reported in percentage, which is impressive to show the effectiveness of our methed.}
\label{table_MMAS}
\resizebox{\linewidth}{!}{
\begin{threeparttable}
\begin{tabular}{ccclllllllll}
\toprule
\multicolumn{3}{c}{\multirow{2.5}{*}{Point Matching}} & \multicolumn{3}{c}{ScanNet: FD@5}                      & \multicolumn{3}{c}{ScanNet: FD@10}                     & \multicolumn{3}{c}{MatterPort3D}                     \\ \cmidrule(r){4-6} \cmidrule(r){7-9} \cmidrule(r){9-12}
\multicolumn{3}{c}{}                     & MMA@1$\uparrow$            & MMA@2$\uparrow$ & MMA@3$\uparrow$ & MMA@1$\uparrow$                    & MMA@2$\uparrow$$\uparrow$            & MMA@3$\uparrow$                      & MMA@1$\uparrow$            & MMA@2$\uparrow$        & MMA@3$\uparrow$               \\ \toprule
\multirow{4}{*}{\rotatebox{90}{Sparse}} & \multicolumn{2}{c}{SP+SG \cite{superglue}} & {37.54}         & 63.06                   & 76.15          & 24.40                   & 43.32          & 57.57       & 13.77    & 21.66      & 29.95  \\
& \multicolumn{2}{c}{GT+SGAM\_SP+SG} & \fs{41.74}$_{+11.18\%}$         & \fs68.31$_{+8.32\%}$                  & \fs81.46$_{+6.98\%}$          & \fs26.44$_{+8.37\%}$                  & \fs44.96$_{+3.79\%}$          & \fs59.60$_{+3.52\%}$       & \fs15.94$_{+15.80\%}$    & \fs24.23$_{+11.87\%}$     & \fs32.86 $_{+9.73\%}$ \\
& \multicolumn{2}{c}{SEEM-L~\cite{SEEM}+SGAM\_SP+SG} & \nd{40.82}$_{+8.73\%}$          & \nd66.68$_{+5.74\%}$                    & \nd80.58$_{+5.82\%}$           & \nd25.65$_{+5.14\%}$                   & \nd44.42$_{+2.54\%}$           & \nd59.42$_{+3.21\%}$        & \nd14.95$_{+8.61\%}$     & \nd23.36$_{+7.85\%}$       & \nd31.96$_{+6.72\%}$   \\
& \multicolumn{2}{c}{SEEM-T~\cite{SEEM}+SGAM\_SP+SG} & 
\rd 39.34$_{+4.79\%}$ & \rd 65.31$_{+3.56\%}$ & \rd78.43$_{+3.00\%}$ & \rd25.31$_{+3.74\%}$ & \rd43.86$_{+1.25\%}$ & \rd58.65$_{+1.87\%}$ & \rd14.14$_{+2.72\%}$ & \rd22.86$_{+5.55\%}$ & \rd31.52$_{+5.25\%}$  \\ \midrule

\multirow{12}{*}{\rotatebox{90}{Semi-Dense}} & \multicolumn{2}{c}{ASpan \cite{ASpan}}                & 32.99          & 66.91                    & 85.03          & 25.35                    & 49.83         & 70.79      & 7.17      &21.10     &37.25   \\
& \multicolumn{2}{c}{GT+SGAM\_ASpan}          & \fs 37.88$_{+14.82\%}$         & \fs 72.81$_{+8.83\%}$    & \fs89.40$_{+5.14\%}$ & \fs28.19$_{+11.19\%}$           & \fs{{54.67}$_{+9.72\%}$}         & \fs{{75.42}$_{+6.53\%}$}         &\fs7.68$_{+7.03\%}$          & \fs{24.51}$_{+16.20\%}$         & \fs{39.94}$_{+7.21\%}$          \\
& \multicolumn{2}{c}{SEEM-L+SGAM\_ASpan}          & \nd 36.48$_{+10.58\%}$         & \nd 70.70$_{+5.66\%}$                & \nd{{87.58}$_{+2.99\%}$}         & \nd{{27.15}$_{+7.09\%}$}                   & \nd{52.85}$_{+6.07\%}$  &\nd73.76$_{+4.19\%}$        & \nd7.61$_{+6.11\%}$   &\nd23.98$_{+13.66\%}$   &\nd39.16$_{+5.12\%}$  \\
& \multicolumn{2}{c}{SEEM-T+SGAM\_ASpan}          & \rd35.54$_{+7.73\%}$         & \rd69.44$_{+3.79\%}$     & \rd86.64$_{+1.89\%}$      & \rd26.81$_{+5.78\%}$     & \rd{{52.11}$_{+4.59\%}$}         & \rd{{72.84}$_{+2.89\%}$}      & \rd7.40$_{+3.17\%}$             & \rd{22.51}$_{+6.69\%}$         & \rd{38.41}$_{+3.11\%}$          \\  \cmidrule(l){2-12}
& \multicolumn{2}{c}{QuadT \cite{QT}}                & 32.79          & 70.40      &88.31       &22.67       & 56.92          & 78.46   & 7.44                 & 23.97         & 41.72                   \\
& \multicolumn{2}{c}{GT+SGAM\_QuadT }          & \fs{39.43}$_{+20.25\%}$          & \fs{75.96}$_{+7.89\%}$    & \fs90.94$_{+2.47\%}$      & \fs26.68$_{+17.65\%}$     & \fs{{62.49}$_{+9.79\%}$}         & \fs{{82.32}$_{+4.93\%}$}               & \fs8.26$_{+11.10\%}$    & \fs{26.19}$_{+9.29\%}$         & \fs{45.56}$_{+9.20\%}$          \\
& \multicolumn{2}{c}{SEEM-L+SGAM\_QuadT }          & \nd{37.02}$_{+12.91\%}$          & \nd{73.63}$_{+4.59\%}$   &\nd89.30$_{+1.12\%}$   &\nd25.17$_{+11.03\%}$         & \nd{{60.55}$_{+6.38\%}$}         & \nd{{81.08}$_{+3.35\%}$}   & \nd7.91$_{+6.41\%}$                & \nd{25.95}$_{+8.26\%}$         & \nd{43.05}$_{+3.17\%}$          \\
& \multicolumn{2}{c}{SEEM-T+SGAM\_QuadT }          & \rd{36.35}$_{+10.85\%}$          & \rd{72.54}$_{+3.04\%}$    & \rd88.40$_{+0.10\%}$   &\rd24.30$_{+7.15\%}$        & \rd{{59.38}$_{+4.32\%}$}         & \rd{{80.46}$_{+2.55\%}$}        & \rd7.86$_{+5.63\%}$           & \rd{24.50}$_{+2.23\%}$         & \rd42.88$_{+2.76\%}$          \\  \cmidrule(l){2-12}
& \multicolumn{2}{c}{LoFTR \cite{loftr}}                & 30.49          & 65.33      &83.51     &17.85      & 46.78          & 67.90   & 9.50                 & 22.08         & 36.07                  \\
& \multicolumn{2}{c}{GT+SGAM\_LoFTR}          & \fs35.02$_{+14.85\%}$          & \fs70.38$_{+7.73\%}$   & \fs88.06$_{+5.45\%}$    & \fs19.02$_{+6.55\%}$         & \fs{{49.10}$_{+4.95\%}$}         & \fs{{70.55}$_{+3.91\%}$}         & \fs12.48$_{+31.36\%}$    &\fs29.08$_{+31.74\%}$      & \fs{48.31}$_{+33.93\%}$               \\
& \multicolumn{2}{c}{SEEM-L+SGAM\_LoFTR}          & \nd33.83$_{+10.95\%}$          & \nd70.05$_{+7.23\%}$                & \nd{{87.33$_{+4.58\%}$}}         & \nd{{18.85$_{+5.60\%}$}}                   & \nd{48.78$_{+4.27\%}$}         & \nd{68.90$_{+1.47\%}$}   & \nd12.27$_{+29.16\%}$  & \nd27.20$_{+23.22\%}$  & \nd40.25$_{+11.57\%}$    \\
& \multicolumn{2}{c}{SEEM-T+SGAM\_LoFTR}           & \rd33.17$_{+10.55\%}$          & \rd69.52$_{+6.40\%}$                & \rd{{86.71$_{+3.84\%}$}}         & \rd{{18.21$_{+2.03\%}$}}                   & \rd{47.45$_{+1.42\%}$}         & \rd{67.98$_{+0.12\%}$}   & \rd11.47$_{+20.77\%}$  & \rd25.10$_{+13.69\%}$  & \rd38.45$_{+6.59\%}$    \\ \midrule
\multirow{4}{*}{\rotatebox{90}{Dense}} & \multicolumn{2}{c}{COTR \cite{cotr}}                 & 32.92          & 63.45                  & 78.71          & 16.51                    & 42.36        & 60.99   & 10.63  & 29.37      &46.07          \\
& \multicolumn{2}{c}{GT+SGAM\_COTR}          & \fs 36.76$_{+11.67\%}$          & \fs 66.56$_{+4.91\%}$   & \fs 81.19$_{+3.16\%}$    & \fs 18.56$_{+12.42\%}$         & \fs{{45.45}$_{+7.28\%}$}         & \fs{{64.52}$_{+5.79\%}$}         & \fs 12.36$_{+16.36\%}$    & \fs 32.64$_{+11.16\%}$      & \fs {49.82}$_{+8.15\%}$               \\
& \multicolumn{2}{c}{{SEEM-L+SGAM\_COTR}}
 & \nd{36.54$_{+11.00\%}$}          & \nd{66.48$_{+4.78\%}$}                & \nd{{81.04$_{+2.96\%}$}}         & \nd{{18.16$_{+10.01\%}$}}                   & \nd{44.54$_{+5.15\%}$}         & \nd{63.29$_{+3.76\%}$}   & \nd{11.73$_{+10.40\%}$}  & \nd{31.97$_{+8.88\%}$}  & \nd{48.70$_{+5.71\%}$}  \\
& \multicolumn{2}{c}{{SEEM-T+SGAM\_COTR}}    & \rd{36.05$_{+9.52\%}$}          & \rd{65.89$_{+3.85\%}$}                & \rd{{80.48$_{+2.24\%}$}}         & \rd{{17.60$_{+6.60\%}$}}                   & \rd{43.71$_{+3.18\%}$}         & \rd{62.50$_{+2.47\%}$}   & \rd{11.11$_{+4.57\%}$}  & \rd{31.30$_{+6.58\%}$}  & \rd{47.49$_{+3.08\%}$}   \\

\bottomrule
\end{tabular}
\end{threeparttable}
}
\vspace{-1.0em}
\end{table*}

\subsection{Dataset}
To demonstrate the superiority of the A2PM framework and SGAM method, we first evaluate our methods on two different indoor datasets, ScanNet \cite{scannet} and MatterPort3D \cite{Matterport3D}. 
Additionally, we investigated the robustness of our method in diverse semantic scenes by conducting experiments on the outdoor KITTI360~\cite{KITTI} and YFCC100M~\cite{yfcc100m} dataset.
First three datasets all offer ground truth semantic labels, which can be directly used as the semantic input of our method.
Moreover, we also evaluated the practicability of our method by utilizing the input from recent semantic segmentation method, SEEM~\cite{SEEM}.
ScanNet contains numerous sequence images, and we selected image pairs with varying levels of difficulty based on the frame difference from its $scene\_0000$ to $scene\_0299$ to evaluate our method.
We also compare with other SOTA methods on the standard ScanNet1500 benchmark~\cite{superglue}, where semantic labels are achieved by SEEM.
Due to the data collection settings of MatterPort3D, image pairs with overlap in this dataset have wide baseline and present challenging matching conditions. These conditions allow us to effectively showcase the performance of our method under difficult matching conditions.
The KITTI360 dataset allows for the evaluation of driving scenes, which is widely-used for SLAM.
The YFCC100M dataset contains internet images of architectural scenarios, which is widely-used for SfM.

\begin{table*}[t]
\caption{\textbf{Relative pose estimation Results (\%).} The AUC of pose error on ScanNet (FD@5/10) and MatterPort3D with different thresholds are reported. \textbf{Our SGAM} is applied on four baselines. {To show the impact of semantic accuracy to our method, we takes three different semantic inputs: \colorbox{colorFst}{SGAM using ground truth (GT)}, \colorbox{colorSnd}{SGAM using SEEM-L} and \colorbox{colorTrd}{SGAM using SEEM-T}.} The improvement achieved by SGAM is also reported in percentage.
}
\label{tab_PAUC_SN}
\resizebox{\textwidth}{!}{ 
\begin{threeparttable}
\begin{tabular}{ccclllllllll}
\toprule
\multicolumn{3}{c}{\multirow{2.5}{*}{Pose Estimation}}                                      & \multicolumn{3}{c}{ScanNet: FD@5}                     & \multicolumn{3}{c}{ScanNet: FD@10}                    & \multicolumn{3}{c}{MatterPort3D}                    \\ \cmidrule(r){4-6} \cmidrule(r){7-9} \cmidrule(r){10-12} 
\multicolumn{3}{c}{} & AUC@$5^{\circ}$$\uparrow$           & AUC@$10^{\circ}$$\uparrow$           & AUC@$20^{\circ}$$\uparrow$           & AUC@$5^{\circ}$$\uparrow$            & AUC@$10^{\circ}$$\uparrow$           & AUC@$20^{\circ}$$\uparrow$           & AUC@$10^{\circ}$$\uparrow$            & AUC@$20^{\circ}$$\uparrow$           & AUC@$30^{\circ}$$\uparrow$           \\ \toprule 
\multirow{4}{*}{\rotatebox{90}{Sparse}} & \multicolumn{2}{c}{SP +SG\cite{superglue}}                  & 67.46          & 76.46          & {86.61}          & 53.11          & 64.47          & 73.58          & 16.39          & 29.54          & 37.61          \\
& \multicolumn{2}{c}{GT+SGAM\_SP+SG }   & \fs 69.20$_{+2.58\%}$ & \fs 78.61$_{+2.81\%}$ & \fs 88.72$_{+2.44\%}$ & \fs 55.87$_{+5.20\%}$ & \fs 66.87$_{+3.72\%}$ & \fs 75.98$_{+3.26\%}$ & \fs 17.93$_{+9.40\%}$ & \fs 31.72$_{+7.38\%}$ & \fs 39.31$_{+4.52\%}$   \\
& \multicolumn{2}{c}{SEEM-L~\cite{SEEM}+SGAM\_SP+SG }   & \nd 68.61$_{+1.70\%}$ & \nd 77.60$_{+1.49\%}$ & \nd 87.44$_{+0.96\%}$ & \nd 53.91$_{+1.51\%}$ & \nd 65.46$_{+1.54\%}$ & \nd 75.03$_{+1.97\%}$ & \nd 17.15$_{+4.64\%}$ & \nd 31.53$_{+6.74\%}$ & \nd 38.51$_{+2.39\%}$  \\
& \multicolumn{2}{c}{SEEM-T~\cite{SEEM}+SGAM\_SP+SG}   & \rd 68.33$_{+1.30\%}$ & \rd 77.35$_{+1.16\%}$ & \rd 87.13$_{+0.60\%}$ & \rd 53.26$_{+0.28\%}$ & \rd 65.11$_{+0.99\%}$ & \rd 74.62$_{+1.41\%}$ & \rd 16.85$_{+2.81\%}$ & \rd 30.37$_{+2.81\%}$ & \rd 37.95$_{+0.90\%}$   \\ \midrule
\multirow{12}{*}{\rotatebox{90}{Semi-Dense}} & \multicolumn{2}{c}{ASpan \cite{ASpan}}         & 70.73         & 77.41          & 80.19          & 58.51 & 70.42          & 79.84          & 18.35  & 27.81          & 43.98          \\
& \multicolumn{2}{c}{GT+SGAM\_ASpan }   & \fs 73.60$_{+4.06\%}$    & \fs 81.52$_{+5.30\%}$ & \fs 85.83$_{+7.02\%}$  & \fs 60.78$_{+3.89\%}$         & \fs 74.24$_{+5.43\%}$ & \fs 84.53$_{+5.87\%}$ & \fs 20.50$_{+11.71\%}$        & \fs 30.08$_{+8.17\%}$ & \fs 48.49$_{+10.26\%}$  \\
& \multicolumn{2}{c}{SEEM-L+SGAM\_ASpan }   & \nd 72.32$_{+2.24\%}$    & \nd 80.59$_{+4.10\%}$ & \nd 85.20$_{+6.24\%}$  & \nd 59.17$_{+1.13\%}$         & \nd 73.02$_{+3.69\%}$ & \nd 83.40$_{+4.46\%}$ & \nd 19.74$_{+7.56\%}$        & \nd 29.38$_{+5.66\%}$ & \nd 45.52$_{+3.51\%}$  \\
& \multicolumn{2}{c}{SEEM-T+SGAM\_ASpan}   & \rd 71.84$_{+1.56\%}$    & \rd 79.94$_{+3.26\%}$ & \rd 83.32$_{+3.90\%}$  & \rd 58.87$_{+0.62\%}$         & \rd 72.02$_{+2.27\%}$ & \rd 82.54$_{+3.38\%}$ & \rd 18.64$_{+1.56\%}$        & \rd 28.87$_{+3.81\%}$ & \rd 44.34$_{+0.83\%}$   \\  \cmidrule(l){2-12}
& \multicolumn{2}{c}{QuadT \cite{QT}}         & 69.48         & 74.25          & 79.39          & 59.27 & 69.77         & 74.96          & 16.53  & 26.98          & 39.96          \\
& \multicolumn{2}{c}{GT+SGAM\_QuadT }   & \fs 72.55$_{+4.41\%}$    & \fs75.89$_{+2.27\%}$ & \fs82.10$_{+3.41\%}$  & \fs 61.82$_{+4.31\%}$         & \fs 71.83$_{+2.95\%}$ & \fs 76.82$_{+2.49\%}$ & \fs18.90$_{+14.33\%}$         & \fs 28.11$_{+4.19\%}$ & \fs 42.21$_{+5.63\%}$   \\

& \multicolumn{2}{c}{SEEM-L+SGAM\_QuadT }   & \nd 71.92$_{+3.51\%}$    & \nd75.60$_{+1.81\%}$ & \nd81.43$_{+2.56\%}$  & \nd 61.78$_{+4.23\%}$         & \nd 71.96$_{+3.14\%}$ & \nd 76.52$_{+2.08\%}$ & \nd17.83$_{+7.82\%}$         & \nd 27.25$_{+1.00\%}$ & \nd 41.56$_{+4.00\%}$  \\
& \multicolumn{2}{c}{SEEM-T+SGAM\_QuadT }   & \rd 71.47$_{+2.86\%}$    & \rd75.06$_{+1.09\%}$ & \rd80.66$_{+1.59\%}$ & \rd 60.97$_{+2.87\%}$         & \rd 71.77$_{+2.86\%}$ & \rd 75.80$_{+1.13\%}$ & \rd16.84$_{+1.88\%}$         & \rd 27.10$_{+0.44\%}$ & \rd 40.22$_{+0.65\%}$   \\  \cmidrule(l){2-12}
& \multicolumn{2}{c}{LoFTR \cite{loftr}}         & 67.69          & 74.29          & 78.45          & 58.71 & 69.81          & 78.99          & 17.98  & 
27.79          & 38.19          \\
& \multicolumn{2}{c}{GT+SGAM\_LoFTR}   & \fs 71.22$_{+5.21\%}$    & \fs79.31$_{+6.76\%}$ & \fs84.44$_{+7.64\%}$  & \fs 59.50$_{+1.35\%}$         & \fs 73.12$_{+4.74\%}$ & \fs 83.13$_{+5.24\%}$ & \fs18.14$_{+0.89\%}$         & \fs 32.28$_{+16.16\%}$ & \fs 45.37$_{+18.80\%}$   \\
& \multicolumn{2}{c}{SEEM-L+SGAM\_LoFTR}   & \nd 70.25$_{+3.79\%}$    & \nd79.23$_{+6.65\%}$ & \nd83.93$_{+6.99\%}$  & \nd 60.37$_{+2.83\%}$         & \nd 71.02$_{+1.74\%}$ & \nd 80.91$_{+2.42\%}$ & \nd18.04$_{+0.31\%}$         & \nd 30.86$_{+11.04\%}$ & \nd 42.13$_{+10.32\%}$   \\
& \multicolumn{2}{c}{SEEM-T+SGAM\_LoFTR}   & \rd 69.53$_{+2.72\%}$    & \rd78.03$_{+5.03\%}$ & \rd82.32$_{+4.93\%}$  & \rd 59.01$_{+0.50\%}$         & \rd 70.58$_{+1.10\%}$ & \rd 79.78$_{+1.01\%}$ & \rd17.47$_{-2.85\%}$         & \rd 30.00$_{+7.94\%}$ & \rd 40.63$_{+6.39\%}$   \\ \midrule
\multirow{4}{*}{\rotatebox{90}{Dense}} &\multicolumn{2}{c}{COTR \cite{cotr}}     & 66.91          & 74.11          & 78.48          & 51.92          & 63.36          & 72.55          & 17.80          & 25.08          & 34.08          \\
& \multicolumn{2}{c}{GT+SGAM\_COTR}    & \fs 71.18$_{+6.38\%}$ & \fs 79.22$_{+6.90\%}$         & \fs 84.22$_{+7.31\%}$          & \fs 53.99$_{+3.99\%}$       & \fs 68.29$_{+7.78\%}$          & \fs  80.17$_{+10.50\%}$          & \fs 19.20$_{+7.87\%}$          & \fs{28.31}$_{+12.88\%}$         & \fs 41.25$_{+21.04\%}$         \\ 
& \multicolumn{2}{c}{SEEM-L+SGAM\_COTR}    & \nd 69.67$_{+4.12\%}$ & \nd78.15$_{+5.46\%}$         & \nd83.84$_{+6.83\%}$         & \nd 52.80$_{+1.70\%}$       & \nd 66.16$_{+4.42\%}$          & \nd 78.78$_{+8.59\%}$          & \nd 18.12$_{+1.82\%}$          & \nd{26.99}$_{+7.62\%}$         & \nd 36.69$_{+7.67\%}$         \\ 
& \multicolumn{2}{c}{SEEM-T+SGAM\_COTR}    & \rd 69.40$_{+3.72\%}$ & \rd77.98$_{+5.32\%}$         & \rd83.07$_{+5.85\%}$          & \rd 52.07$_{+0.28\%}$        & \rd 65.91$_{+4.03\%}$          & \rd 78.43$_{+8.10\%}$          & \rd 17.93$_{+0.71\%}$          & \rd{25.79}$_{+2.84\%}$         & \rd 35.79$_{+5.01\%}$        \\  
\bottomrule
\end{tabular}
\end{threeparttable}
}
\end{table*}

\subsection{Point Matching}
We first conduct point matching experiments on ScanNet and MatterPort3D. For ScanNet, we construct two matching difficulties with image pairs under various \textit{Frame Differences} (FD@5/10), each including 1500 image pairs. 
For the more challenging condition, 500 image pairs are sampled from the first 5 scenes in MatterPort3D.

\textit{1) Compared methods.} We compare the proposed SGAM method with various matching methods of all three types, including sparse method~\cite{superpoint}, semi-dense methods~\cite{ASpan,QT,loftr} and dense method~\cite{cotr}.
Particularly, we combine SGAM with SOTA \textbf{sparse} method (SGAM\_SP+SG~\cite{superpoint,superglue}), \textbf{semi-dense} methods (SGAM\_ASpan~\cite{ASpan}, SGAM\_QuadT~\cite{QT} and {SGAM\_LoFTR~\cite{loftr}}) and \textbf{dense} methods (SGAM\_COTR~\cite{cotr}), to demonstrate the improvement we achieved.
To evaluate the robustness against semantic input, we offers three semantic segmentation sources for SGAM, \ie the ground truth label (GT) and semantic segmentation method SEEM \cite{SEEM} with two backbones: {the \textbf{Large} FocalNet \cite{focal} (SEEM-L) and the \textbf{Tiny} one (SEEM-T). SEEM-L is more accurate than SEEM-T in semantic segmentation.} The implementation details of our SGAM can be found in Sec.~\ref{sec:imp_de} of the appendix.

\textit{2) Evaluation protocol.}  Following \cite{d2net,alike}, we report the Mean Matching Accuracy (MMA@i) in percentage under integer thresholds $i\in [1,3]$ of each method. This metric indicates the proportion of correct matches among all matches.
The number of matches is set as 500 for each method.


\textit{3) Results.} The MMA values are reported in Tab.~\ref{table_MMAS}, with improvement achieved by our methods are shown in percentage. 
It is evident that the SGAM significantly enhances the matching precision of \textbf{all three kinds} of matching methods on \textbf{both} datasets, highlighting the robustness and effectiveness of our approach.
Specifically, although ScanNet serves as the training dataset for ASpan, QuadT, and LoFTR, SGAM still demonstrates impressive accuracy improvement for these methods.
On MatterPort3D dataset, SGAM also exhibits substantial precision improvement, thus underscoring the superiority of SGAM in challenging matching scenarios. 
For each baseline, Tab. \ref{table_MMAS} also presents a comparison of different semantic inputs of our method.
The ground truth labels are most precise, while SEEM, trained with COCO \cite{COCO} labels, may introduce unidentified objects, resulting in a slight decrease in precision.
SGAM using SEEM-L gets higher accuracy than using SEEM-T, demonstrating the matching accuracy increases with the semantic segmentation precision.
Nevertheless, the semantic prior provided by the current semantic segmentation method is sufficient for our approach to achieve remarkable accuracy improvements in matching tasks.
In sum, owing to our improved search space, which alleviates many matching challenges and provides high-resolution input for point matchers, SGAM notably improves the matching accuracy. We offer the visualization in Fig.~\ref{fig_qrsn} of the appendix.

\begin{table*}[t]
\centering
\caption{\textbf{Relative pose estimation Results (\%) on KITTI360 dataset.} {We compare two different semantic inputs for \textbf{our method}: \colorbox{colorFst}{SGAM using ground truth (GT)} and \colorbox{colorSnd}{SGAM using SEEM-L}.}}
\label{tab_PAUC_KITTI}
\resizebox{\linewidth}{!}{
\begin{tabular}{llllllllllll}
\toprule
\multicolumn{3}{c}{\multirow{2.5}{*}{Pose Estimation}}                                      & \multicolumn{3}{c}{Sqe. 00}                     & \multicolumn{3}{c}{Seq. 03}                    & \multicolumn{3}{c}{Seq. 05}                    \\ \cmidrule(r){4-6} \cmidrule(r){7-9} \cmidrule(r){10-12} 
\multicolumn{3}{c}{} & AUC@$5^{\circ}$$\uparrow$           & AUC@$10^{\circ}$$\uparrow$          & AUC@$20^{\circ}$$\uparrow$           & AUC@$5^{\circ}$$\uparrow$            & AUC@$10^{\circ}$           & AUC@$20^{\circ}$$\uparrow$           & AUC@$10^{\circ}$$\uparrow$            & AUC@$20^{\circ}$$\uparrow$           & AUC@$30^{\circ}$$\uparrow$           \\ \midrule
\multirow{4}{*}{\rotatebox{90}{Sparse}} & \multicolumn{2}{c}{SFD2~\cite{sfd2}}                   & 68.58       & 83.24       & 92.30 & 72.71 & 88.04 & 94.31 & 63.18 & 80.52 & 90.95       \\ 
& \multicolumn{2}{c}{SP~\cite{superpoint}+SG~\cite{superglue}}                   & 69.65       & 85.44       & 93.91 & 74.24 & 89.15 & 96.03 & 63.91 & 81.34 & 91.34       \\
 & \multicolumn{2}{c}{GT+SGAM\_SP+SG}        & \fs 72.17$_{+3.62\%}$ & \fs 86.31$_{+1.02\%}$ & \fs 94.37$_{+0.49\%}$ & \fs 75.36$_{+1.51\%}$ & \fs 90.10$_{+1.07\%}$ & \fs 97.26$_{+1.28\%}$ & \fs 65.01$_{+1.72\%}$ & \fs 82.78$_{+1.77\%}$ & \fs 92.61$_{+1.39\%}$     \\ 
& \multicolumn{2}{c}{SEEM-L\cite{SEEM}+SGAM\_SP+SG}        & \nd 71.93$_{+3.27\%}$ & \nd 86.06$_{+0.73\%}$ & \nd 94.11$_{+0.21\%}$ & \nd 75.23$_{+1.33\%}$ & \nd 89.87$_{+0.81\%}$ & \nd 96.99$_{+1.00\%}$ & \nd 64.85$_{+1.47\%}$ & \nd 82.53$_{+1.46\%}$ & \nd 92.17$_{+0.91\%}$     \\ \midrule
\multirow{9}{*}{\rotatebox{90}{Semi-Dense}} &\multicolumn{2}{c}{ASpan\cite{ASpan}}                   & 61.19       & 77.64       & 87.95 & 68.01 & 83.00 & 91.20 & 57.38 & 75.57 & 87.16       \\
& \multicolumn{2}{c}{GT+SGAM\_ASpan}        & \fs {66.05}$_{+7.95\%}$      & \fs {{81.86}}$_{+5.44\%}$      & \fs {90.63}$_{+3.05\%}$ & \fs 73.81$_{+8.53\%}$ & \fs 86.43$_{+4.14\%}$ & \fs 92.96$_{+1.94\%}$ & \fs 63.31$_{+10.35\%}$ & \fs 80.22$_{+6.15\%}$ & \fs 89.78$_{+3.00\%}$     \\ 
& \multicolumn{2}{c}{SEEM-L\cite{SEEM}+SGAM\_ASpan}        & \nd {66.18}$_{+8.17\%}$      & \nd {{81.67}}$_{+5.19\%}$      & \nd {90.38}$_{+2.76\%}$ &   \nd 73.76$_{+8.46\%}$ & \nd 86.40$_{+4.10\%}$ & \nd 92.95$_{+1.92\%}$ & \nd 63.06$_{+9.91\%}$ & \nd 80.07$_{+5.95\%}$ & \nd 89.70$_{+2.92\%}$     \\  \cmidrule(l){2-12}
& \multicolumn{2}{c}{QuadT\cite{QT}}                   & 59.93       & 77.77       & 88.27      & 66.47 & 81.81 & 90.39 & 58.93 & 77.24 & 88.40 \\
& \multicolumn{2}{c}{GT+SGAM\_QuadT}        & \fs 67.77$_{+13.10\%}$       & \fs 82.68$_{+6.32\%}$       & \fs 91.04$_{+3.13\%}$  & \fs 73.40$_{+10.42\%}$ & \fs 86.05$_{+5.18\%}$ & \fs 92.68$_{+2.53\%}$ & \fs 65.98$_{+11.96\%}$ & \fs 81.97$_{+6.11\%}$ & \fs 90.83$_{+2.75\%}$  
     \\
& \multicolumn{2}{c}{SEEM-L\cite{SEEM}+SGAM\_QuadT}        & \nd 67.72$_{+13.01\%}$       & \nd 82.63$_{+6.25\%}$       & \nd 91.01$_{+3.10\%}$    & \nd 73.33$_{+10.32\%}$ & \nd 86.00$_{+5.12\%}$ & \nd 92.65$_{+2.51\%}$ & \nd 65.92$_{+11.86\%}$ & \nd 81.77$_{+5.86\%}$ & \nd 90.67$_{+2.57\%}$    \\ \cmidrule(l){2-12}
& \multicolumn{2}{c}{LoFTR\cite{loftr}}                   & 65.11       & 80.98       & 90.10    & 71.53 & 84.87 & 92.11 & 63.54 & 80.19 & 89.95   \\
& \multicolumn{2}{c}{GT+SGAM\_LoFTR}        & \fs 70.55$_{+8.36\%}$       & \fs 84.79$_{+4.71\%}$       & \fs 92.33$_{+2.48\%}$  & \fs 75.40$_{+5.41\%}$  & \fs 86.95$_{+2.45\%}$ & \fs 93.03$_{+1.00\%}$ & \fs 69.68$_{+9.66\%}$ & \fs 84.44$_{+5.30\%}$ & \fs 92.20$_{+2.50\%}$   \\
& \multicolumn{2}{c}{SEEM-L\cite{SEEM}+SGAM\_LoFTR}        & \nd 70.20$_{+7.83\%}$       & \nd 84.54$_{+4.39\%}$       & \nd 92.16$_{+2.29\%}$  & \nd 76.21$_{+6.54\%}$ & \nd 87.27$_{+2.83\%}$ & \nd 93.18$_{+1.16\%}$ & \nd 69.51$_{+9.39\%}$ & \nd 84.11$_{+4.89\%}$ & \nd 92.01$_{+2.29\%}$  \\ \midrule
\multirow{3}{*}{\rotatebox{90}{Dense}} & \multicolumn{2}{c}{COTR\cite{cotr}}                   & 62.76       & 77.61       & 86.67  & 66.97 & 80.92 & 89.22 & 58.69 & 79.36 & 88.55    \\
& \multicolumn{2}{c}{GT+SGAM\_COTR}        & \fs 67.55$_{+7.63\%}$       & \fs 81.70$_{+5.27\%}$       & \fs 88.30$_{+1.89\%}$ & \fs 72.40$_{+8.11\%}$ & \fs 85.89$_{+6.15\%}$  & \fs 91.32$_{+2.35\%}$ & \fs 66.17$_{+12.75\%}$ & \fs 82.01$_{+3.34\%}$ & \fs 90.96$_{+2.72\%}$  
  \\
& \multicolumn{2}{c}{SEEM-L\cite{SEEM}+SGAM\_COTR}        & \nd 66.69$_{+6.26\%}$       & \nd 80.90$_{+4.23\%}$       & \nd 88.37$_{+1.96\%}$  & \nd 70.17$_{+4.77\%}$ & \nd 84.21$_{+4.07\%}$ & \nd 90.17$_{+1.06\%}$ & \nd 64.39$_{+9.72\%}$ & \nd 81.19$_{+2.31\%}$ & \nd 90.18$_{+1.85\%}$    \\

\bottomrule
\end{tabular}
}
\end{table*}


\begin{table}[t]
\centering
\caption{\textbf{Relative pose estimation Results (\%) on ScanNet1500 benchmark.}  {\colorbox{colorFst}{Our method} obtains the semantic prior by SEEM-L.} The \textbf{best} and \uline{second} results are highlighted.}
\label{tab_PAUC_ST}
\resizebox{\linewidth}{!}{
\begin{threeparttable}
\begin{tabular}{lllllll}
\toprule
\multicolumn{3}{l}{\multirow{2}{*}{Pose Estimation}} & \multicolumn{3}{c}{ScanNet1500 benchmark} \\ \cmidrule(l){4-6} 
\multicolumn{3}{l}{}                        & AUC@5$^\circ$$\uparrow$       & AUC@10$^\circ$$\uparrow$       & AUC@20$^\circ$$\uparrow$        \\ \midrule

&  \multicolumn{2}{l}{PATS\cite{pats}}                    & {26.00}       & {46.90}       & {64.30}       \\ 
\midrule
\multirow{4}{*}{\rotatebox{90}{Sparse}} &  \multicolumn{2}{l}{SP\cite{superpoint}+OANet\cite{OANet}}                & 11.80       & 26.90       & 43.90       \\
&\multicolumn{2}{l}{SP+SG\cite{superglue}}                   & 16.20       & 33.80       & 51.80       \\ 
& \multicolumn{2}{l}{{MKPC~\cite{MKPC}+SP+SG}}                    & {{16.18}$_{-0.12\%}$}       & {{34.11}$_{+0.92\%}$}       & {{52.47}$_{+1.29\%}$ }
\\
& \multicolumn{2}{l}{{SEEM-L~\cite{SEEM}+SGAM\_SP+SG}}                    & \fs{{17.33}$_{+6.98\%}$}       & \fs{{34.77}$_{+2.87\%}$}       & \fs{{52.13}$_{+0.64\%}$ }

\\
\midrule
\multirow{6}{*}{\rotatebox{90}{Semi-Dense}} &\multicolumn{2}{l}{ASpan~\cite{ASpan}}                   & 25.78       & 46.14       & 63.32       \\
& \multicolumn{2}{l}{SEEM-L+SGAM\_ASpan}        & \fs{27.51}$_{+6.71\%}$      & \fs{{48.01}}$_{+4.05\%}$      & \fs{65.26}$_{+3.06\%}$       \\   \cmidrule(l){2-6}
& \multicolumn{2}{l}{QuadT~\cite{QT}}                   & 25.21       & 44.85       & 61.70       \\
& \multicolumn{2}{l}{SEEM-L+SGAM\_QuadT}        & \fs25.53$_{+1.27\%}$       & \fs46.02$_{+2.60\%}$       & \fs63.40$_{+2.76\%}$       \\  \cmidrule(l){2-6}
& \multicolumn{2}{l}{LoFTR~\cite{loftr}}                   & 22.13       & 40.86       & 57.65       \\
& \multicolumn{2}{l}{ SEEM-L+SGAM\_LoFTR}        & \fs23.39$_{+5.69\%}$       & \fs41.79$_{+2.28\%}$       & \fs58.74$_{+1.89\%}$       \\ \midrule
\multirow{2}{*}{\rotatebox{90}{Dense}} &\multicolumn{2}{l}{{DKM~\cite{dkm}}}                   & {\uline{29.40}}       & {\uline{50.74}}       & {\uline{68.31}}       \\[0.5ex]
& \multicolumn{2}{l}{{SEEM-L+SGAM\_DKM}}        & \fs{\textbf{30.61}$_{+4.12\%}$}       & \fs{\textbf{52.34}$_{+3.10\%}$ }      & \fs{\textbf{69.31}$_{+1.48\%}$}       \\[0.5ex]
\bottomrule
\end{tabular}
\end{threeparttable}
}
\end{table}

\subsection{Relative Pose Estimation}
Accurate point matches do not necessarily lead to accurate geometry, as point distribution is also important. 
Thus we evaluate our method for relative pose estimation.
The dataset used in our evaluation comprises both indoor and outdoor scenes. Specifically, we employ ScanNet and Matterport3D for indoor scenes.
We sample $2\!\times\!1500$ image pairs from ScanNet (FD@5/10) and 500 image pairs from MatterPort3D to construct three difficulties.
We also investigate the influence of different semantic inputs.
Additionally, we compare our method with more SOTA methods \cite{pats,OANet,MKPC} on the standard ScanNet1500 benchmark \cite{superglue}, using {SEEM-L} to obtain the semantic segmentation results.
For outdoor scenes, we use the KITTI360~\cite{KITTI} and YFCC100M~\cite{yfcc100m} dataset.
{Due to the \textit{static world assumption}~\cite{SLAMSurvey,COLMAP} in downstream tasks, we utilized three sequences in the KITTI360 dataset (Seq. 00, 03 and 05) with few moving objects, \eg pedestrians, for pose estimation estimation.
In these sequences, we showcase the improvement achieved by our method across five baselines, using the semantic prior from both ground truth and {SEEM-L} results.
For YFCC100M dataset, we follow the previous work~\cite{superglue} to construct pose estimation evaluation with 4k image pairs. We also use SEEM-L and SEEM-T to obtain semantic prior of this dataset for our method.}

\begin{figure*}[!t]
\centering
\includegraphics[width=\linewidth]{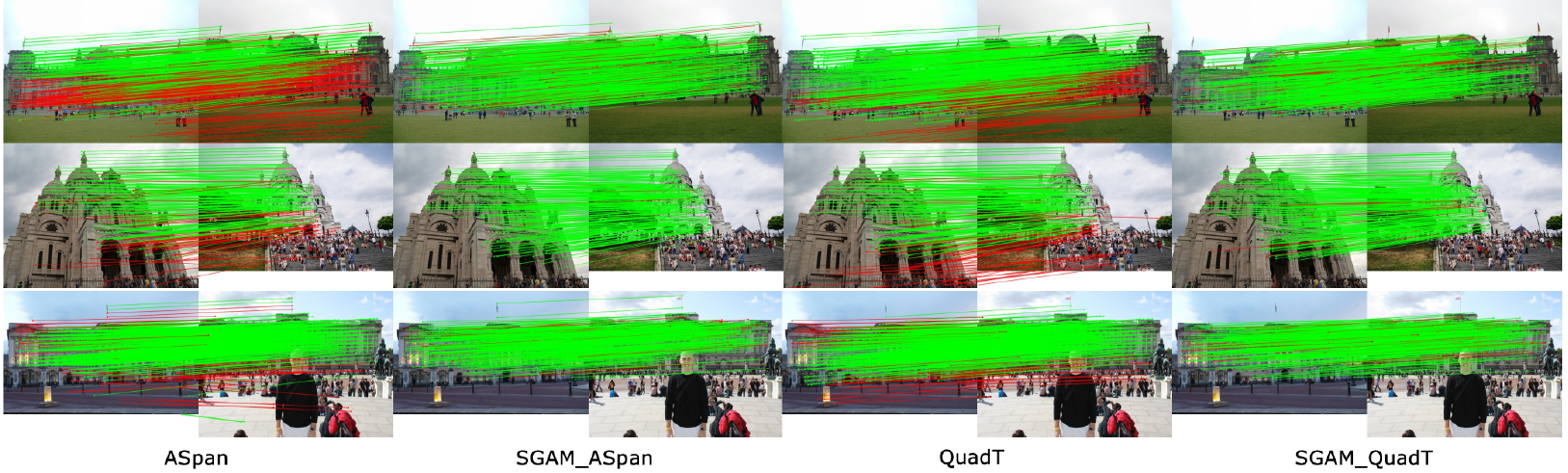}
\caption{{\textbf{Qualitative Comparison on YFCC100M.} The visual comparison between our method and two SOTA baselines.
The \textcolor{red}{wrong} and \textcolor{green}{correct} matches under the same threshold are labeled respectively.
} }
\vspace{-1.2em}
\label{fig_qrYFCC}
\end{figure*}

\textit{1) Evaluation protocol.} Following \cite{loftr}, we report the pose estimation AUC, which reveals the proportion of correct pose estimation among all estimations.
The camera pose is recovered by solving the essential matrix with RANSAC. {Correspondences are uniformly sampled from the image, with a maximum number of 500.}
SGAM is also combined with three kinds of point matchers to demonstrate the advantages adopting the proposed search space. We replace COTR with DKM~\cite{dkm} as the proxy of dense methods, due to its impressive performance.
For ScanNet, KITTI360 and YFCC100M dataset, we report the pose AUC$@5^{\circ}/10^{\circ}/20^{\circ}$. 
As pose estimation is hard in MatterPort3D, we report the pose AUC$@10^{\circ}/20^{\circ}/30^{\circ}$. 

\begin{table}[t]
\centering
{\revised
\caption{{\textbf{Relative pose estimation Results (\%) on YFCC100M.} Two different semantic inputs for our method are compared: \colorbox{colorFst}{SGAM using SEEM-L} and \colorbox{colorSnd}{SGAM using SEEM-T}.}}
\label{tab_PAUC_YFCC}
\resizebox{\linewidth}{!}{
\begin{threeparttable}
\begin{tabular}{lllll}
\toprule
\multicolumn{2}{c}{\multirow{2}{*}{Pose Estimation}} & \multicolumn{3}{c}{YFCC100M} \\ \cmidrule(l){3-5} 
 & & AUC@5$^\circ$$\uparrow$  & AUC@10$^\circ$$\uparrow$ & AUC@20$^\circ$$\uparrow$ \\ \midrule
&  {PATS\cite{pats}}                    & {39.25}       & {60.77}       & {76.38}       \\ \midrule
\multirow{5}{*}{\rotatebox{90}{Sparse}} & SP+OANet~\cite{OANet}                 & 26.82  & 45.04  & 62.17  \\
 & SP+SG~\cite{superglue}                    & 28.45  & 48.60   & 67.19  \\
& OETR~\cite{OETR}+SP+SG               & 31.51$_{+10.76\%}$ & 50.61$_{+4.14\%}$ & 70.02$_{+4.21\%}$  \\
&  SEEM-L~\cite{SEEM}+SGAM\_SP+SG              & \fs29.54$_{+3.83\%}$ & \fs50.48$_{+3.87\%}$ & \fs69.64$_{+3.65\%}$  \\
& SEEM-T~\cite{SEEM}+SGAM\_SP+SG              & \nd29.14$_{+2.43\%}$ & \nd50.01$_{+2.90\%}$ & \nd68.26$_{+1.59\%}$  \\
\midrule
\multirow{12}{*}{\rotatebox{90}{Semi-Dense}} & ASpan~\cite{ASpan}                    & 38.96  & 59.35  & 75.54  \\
& OETR+ASpan               & 39.31$_{+0.90\%}$ & 60.13$_{+1.31\%}$ & 76.22$_{+0.90\%}$  \\ 
& SEEM-L+SGAM\_ASpan        & \fs39.90$_{+2.41\%}$  & \fs60.36$_{+1.70\%}$  & \fs76.34$_{+1.06\%}$   \\ 
& SEEM-T+SGAM\_ASpan        & \nd39.77$_{+2.08\%}$  & \nd60.24$_{+1.50\%}$  & \nd76.21$_{+0.89\%}$   \\  \cmidrule(l){2-5}
& QuadT~\cite{QT}                       & 40.73  & 61.19  & 76.57  \\
& OETR+QuadT               & 41.46$_{+1.79\%}$ & 62.15$_{+1.57\%}$ & 77.08$_{+0.67\%}$  \\ 			
&  SEEM-L+SGAM\_QuadT           & \fs41.32$_{+1.45\%}$  & \fs61.33$_{+0.23\%}$  & \fs76.79$_{+0.29\%}$  \\ 
& SEEM-T+SGAM\_QuadT           & \nd41.07$_{+0.83\%}$  & \nd61.44$_{+0.41\%}$  & \nd77.02$_{+0.58\%}$  \\ \cmidrule(l){2-5}
& LoFTR~\cite{loftr}                    & 41.12  & 61.43  & 77.01  \\
& OETR+LoFTR               & 41.83$_{+1.73\%}$ & 62.16$_{+1.19\%}$ & 77.35$_{+0.44\%}$  \\ 		
&  SEEM-L+SGAM\_LoFTR        & \fs{41.54}$_{+0.95\%}$  & \fs{61.72}$_{+0.47\%}$  & \fs{77.12}$_{+0.14\%}$  \\
&  SEEM-T+SGAM\_LoFTR        & \nd{41.33}$_{+0.51\%}$  & \nd{61.67}$_{+0.39\%}$   & \nd{77.08}$_{+0.09\%}$  \\ 
\midrule
\multirow{4}{*}{\rotatebox{90}{Dense}} &DKM~\cite{dkm}                       & 43.12  & 63.78  & 79.13  \\ 
& OETR+DKM               & 43.28$_{+0.37\%}$ & 64.27$_{+0.77\%}$ & 79.34$_{+0.27\%}$  \\  		
&  SEEM-L+SGAM\_DKM           & \fs43.77$_{+1.51\%}$  & \fs64.12$_{+0.53\%}$  & \fs79.94$_{+1.01\%}$  \\
&  SEEM-T+SGAM\_DKM           & \nd43.56$_{+1.02\%}$  & \nd63.99$_{+0.33\%}$  & \nd79.77$_{+0.81\%}$  \\
\bottomrule
\end{tabular}
\end{threeparttable}
}
}
\end{table}

\textit{2) Indoor Results.} 
The pose AUC results in indoor scene are summarised in Tab.~\ref{tab_PAUC_SN} and Tab.~\ref{tab_PAUC_ST}. 
In Tab.~\ref{tab_PAUC_SN}, it can be seen that our method consistently improves performance \textbf{across all point matching baselines}, indicating the versatility of our approach.
The impressive precision improvement achieved on the challenging MatterPort3D dataset underscores the ability of our method in tackling difficult matching scenarios.
Furthermore, similar to the point matching experiment, higher semantic precision leads to improved geometry estimation.
The recent SEEM is able to offer enough accurate semantic prior for our method, even hand-crafted feature-based method is applied in SGAM.
Additionally, in Table~\ref{tab_PAUC_ST}, we compare SGAM with other leading approaches on ScanNet1500 benchmark. The results demonstrate that our method boosts all the baselines by a considerable margin, achieving the highest accuracy. {Meanwhile, the co-visible area estimation method, MKPC~\cite{MKPC}, achieves very limited improvement for the sparse baseline, compared to ours (\eg $16.18_{-0.12\%}$ vs. $17.33_{+6.98\%}$). This may be interpreted as MKPC can only achieve single area match for each pair of images. Moreover, its area matching accuracy is dependent on precise point matching, which is challenging to achieve in complex indoor scenes. In contrast, SGAM is particularly well-suited for indoor scenes due to the abundance of semantic information, making it more effective to reduce redundant computation and improve the accuracy.
} 

\textit{3) Outdoor Results.} The results on KITTI360 dataset are reported in Tab.~\ref{tab_PAUC_KITTI}.
Our method greatly enhances the precision of pose estimation for all baselines, affirming the effectiveness and robustness of our method in outdoor driving scenes.
It is important to highlight that the impact of different semantic inputs on performance is minimal here. This is primarily due to the reduced semantic complexity of driving scenes compared to indoor scenes, wherein the {SEEM-L} backbone can yield accurate semantic segmentation outcomes. 
The table also includes a comparison with SFD2~\cite{sfd2}, which incorporates semantic perception trained specifically for driving scenes. While the integration of semantics in SFD2 benefits matching in difficult scenes~\cite{sfd2}, enhancing features of detailed search spaces by semantic remains a challenge.
Conversely, SGAM exhibits superior performance, underscoring the effectiveness of our semantic-friendly search space.

The Tab.~\ref{tab_PAUC_YFCC} reports the results on YFCC100M dataset.
As we can seen in the table, SGAM is able to boost the performance for all baselines, including sparse, semi-dense and dense methods. However, the improvements in this outdoor scene are somewhat restricted (up to $+2.41\%$), compared to indoor and driving scenes. This limitation arises from the scarcity of semantic information in the scene, leading to semantic segmentation at a broad granularity level, e.g. segmentation only contains labels like `sky', `building' and `people'.
Consequently, SGAM generates few area matches, typically encompassing almost the entire image.
Conversely, the co-visible matching method, OETR, can match co-visible areas without considering semantics. Therefore, OETR can reduce the redundant computation more effectively and surpass SGAM in terms of accuracy for sparse baseline. Despite that, SGAM demonstrates a comparable improvement against OETR for semi-dense and dense baselines, suggesting that the high accuracy of baselines may compensate for shortcomings in area matching accuracy.
Additionally, we provide some qualitative comparison examples in Fig.~\ref{fig_qrYFCC}.

\begin{table*}[t]
\centering
\caption{\textbf{Area matching performance on ScanNet.} The area matching results (\%) of SAM and SGAM combined with different point matchers under three matching difficulties {and three semantic input settings} in ScanNet are reported. {The \textbf{best} and \uline{second} results under each semantic input setting are highlighted.} }
\label{table_amp}
 \resizebox{\linewidth}{!}{
\begin{tabular}{ccccccccccccccccc}
\toprule
\multicolumn{2}{c}{Semantic}         & \multicolumn{5}{c}{GT}                                          & \multicolumn{5}{c}{{SEEM-L}}  & \multicolumn{5}{c}{{SEEM-T}}                                       \\ \cmidrule(r){1-2} \cmidrule(r){3-7} \cmidrule(r){8-12} \cmidrule(r){13-17} 
\multicolumn{2}{c}{Method}    & SAM   & \multicolumn{4}{c}{SGAM} & {SAM}   & \multicolumn{4}{c}{{SGAM}} & {SAM}   & \multicolumn{4}{c}{{SGAM}} \\ \cmidrule(r){1-2} \cmidrule(r){3-7} \cmidrule(r){8-12} \cmidrule(r){13-17}  
\multicolumn{2}{c}{Point Matcher}    & -   & ASpan & QuadT    & LoFTR & COTR & {-}  & {ASpan} & {QuadT}    & {LoFTR} & {COTR} & {-}  & {ASpan} & {QuadT}    & {LoFTR} & {COTR} \\ \cmidrule(r){1-2} \cmidrule(r){3-7} \cmidrule(r){8-12} \cmidrule(r){13-17} 
\multirow{2}{*}{FD@5} & AOR$\uparrow$ & 84.95 & 89.54       & \textbf{91.40} & \uline{90.03}       & 89.41      & {85.12} & {85.27}       & {\textbf{86.18}} & {\uline{86.14}}       & {85.31} & {78.70} & {79.30}       & {\uline{80.43}} & {\textbf{80.69}}       & {79.34}        \\
& AMP@0.7$\uparrow$ & 89.42 & 93.43       & \textbf{98.45} & \uline{94.35}       & 92.43      & {87.26} & {\textbf{96.43}}       & {\uline{89.28}} & {88.98}       & {87.47}  & {77.44} & {79.27}       & {\uline{80.32}} & {\textbf{80.44}}       & {80.01}     \\ \cmidrule(r){1-2} \cmidrule(r){3-7} \cmidrule(r){8-12} \cmidrule(r){13-17} 
\multirow{2}{*}{FD@10} & AOR$\uparrow$ & 84.38 & 84.52       & \textbf{87.69} & \uline{85.46}       & 84.62      & {80.08} & {80.16}       & {\textbf{82.34}} & {\uline{81.28}}       & {80.93} & {72.84} & {\uline{74.38}}       & {\textbf{74.80}} & {\textbf{74.80}}       & {73.49}       \\
& AMP@0.7$\uparrow$ & 88.71 & 92.87       & \textbf{97.57} & \uline{93.07}       & 89.25      & {79.07} & {\textbf{85.52}}       & {\uline{82.08}} & {81.51}       & {80.52}  & {67.28} & {68.95}       & {69.16} & {\textbf{69.84}}       & {\uline{69.30}}   \\ \cmidrule(r){1-2} \cmidrule(r){3-7} \cmidrule(r){8-12} \cmidrule(r){13-17} 
\multirow{2}{*}{FD@30} & AOR$\uparrow$ & 69.46 & 72.29       & \textbf{79.95} & \uline{72.56}       & 70.03      & {69.97} & {71.75}       & {\uline{72.80}} & {\textbf{72.95}}       & {70.23}   & {61.84} & {64.85}       & {\uline{65.37}} & {\textbf{65.82}}       & {65.34}     \\
& AMP@0.7$\uparrow$ & 75.32 & 80.56       & \textbf{88.41} & \uline{82.15}       & 76.72      & {58.52} & {\textbf{64.05}}       & {62.68} & {\uline{63.04}}       & {60.51}  & {45.49} & {47.72} & {\uline{48.51}} & {\textbf{48.91}} & {47.34}   \\ \bottomrule
\end{tabular}
}
\end{table*}

\subsection{Area Matching}
We also evaluate SGAM on ScanNet dataset \cite{scannet} for area matching performance. 
{
We sample 3$\times$1500 image pairs for three matching difficulties (FD@5/10/30) in ScanNet.
The impact of semantic precision for area matching is investigated using three semantic input (GT and SEEM-L/T).
}
Furthermore, we compare the performance of SAM alone with that of SGAM integrated with different point matchers.

\textit{1) Evaluation protocol.} 
To measure the area matching accuracy, we propose two area matching metrics as follows. 

\textbf{(a) Area Overlap Ratio (AOR).} 
This metric is to evaluate the single area match accuracy and achieved by projecting points ($\{p_i\}_i^N $) of $\alpha \in I_0$ to $I_1$ and getting the proportion of points falling into the matched area $\beta \in I_1$.
\begin{equation}
    AOR(A) = \frac{1}{N}\sum_i^N(C(P(p_i), \beta))
\end{equation}
where the area match $A=(\alpha, \beta)$, $P(p_i)$ is projecting point $p_i$ to $I_1$, $C(q_i, \beta)$ is 1 when $q_i \in \beta$, otherwise $0$.

\textbf{(b) Area Matching Precision@t (AMP@t).} 
Given all area matches $\{\mathcal{A}_{i,\pi(i)}\}_i^M$ and a specific threshold $t \in [0,1]$, this metric is the proportion of area matches whose AOR $> t$, evaluating the overall matching accuracy.
\begin{equation}
    AMP@t = \frac{1}{M} \sum_i^M F(\mathcal{A}_{i,\pi(i)}, t)
\end{equation}
where {\small$F(\mathcal{A}_{i,\pi(i)}, t)$} is 1 when {\small$AOR(\mathcal{A}_{i,\pi(i)}) \! > \!t$}, otherwise $0$.

\begin{figure}[!t]
\centering
\includegraphics[width=\linewidth]{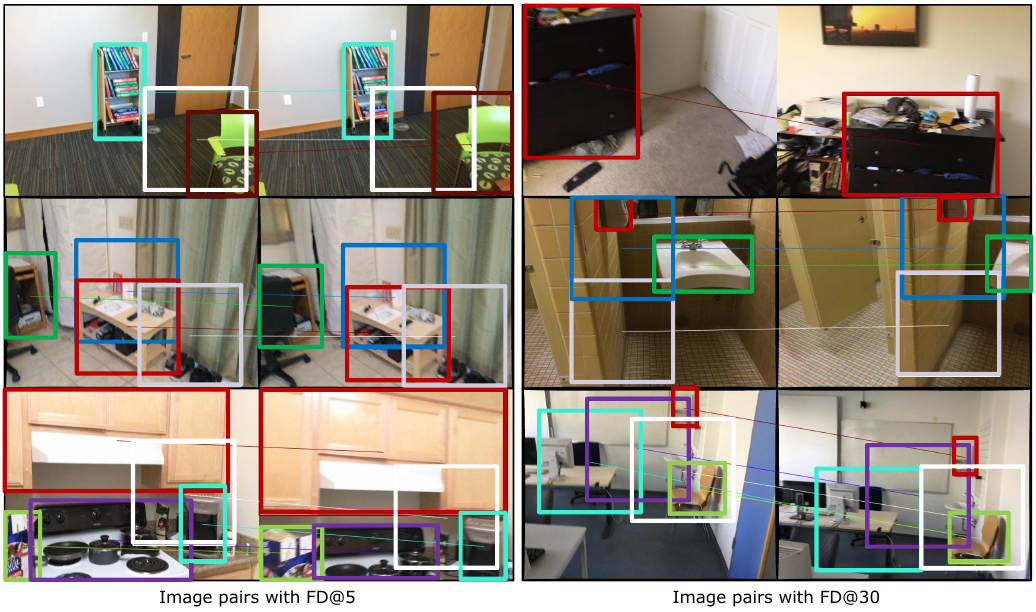}
\caption{\textbf{Qualitative Results of Area Matching.} We show the area matching results of SGAM on ScanNet dataset, {using image pairs with two frame differences (FD@5 and FD@30). Each area match is indicated by a box pair with the same color.} Two kinds of semantic areas can be seen in these cases, i.e. the semantic object areas {centered in objects} and the semantic intersection areas {between objects}, covering most of the overlap.}
\vspace{-1.2em}
\label{fig_AMQR}
\end{figure}

\textit{2) Results.} 
The area matching results are summarised in Tab. \ref{table_amp}. 
{
The threshold $t$ of AMP is set as $0.7$.} 
We analyze the outcomes of SAM and SGAM when combined with SOTA detector-free matchers.
{
Within the table, the precision of area matching in SAM decreases as the matching difficulty increases. When the semantic input is accurate (GT), the AMP values demonstrate that most areas are accurately matched under all conditions.
However, as semantic precision decreases, our method also experiences a decrease in precision, which is more pronounced in large FD. This demonstrates the main limitation of our method, i.e. the heavy reliance on semantic, which is discussed in detail in Sec.~\ref{sec:al}.}
Notably, the utilization of SGAM enhances the accuracy of area matching in all scenarios, highlighting the importance of GAM in area matching.
{Different point matchers also result in different regional matching precision, but the overall difference is small, proving the compatibility of our method for point matchers.}

\subsection{Understanding SAM}
SAM contains matching two kinds of areas: the semantic object area (SOA) and the semantic intersection area (SIA).
To assess the importance of these two areas on area matching, we designed experiments to evaluate their quantities and matching accuracy in ScanNet image pairs with FD@5/10/30.
{
To further investigate the performance under different semantic input accuracy, SAM takes two semantic segmentation as input, including ground truth (\textbf{GT sem.}) and segmentation by SEEM with large backbones (\textbf{SEEM-L Sem.}).
The results are summarised in the Tab.~\ref{tab_SAM_GP_ab}, including the \textbf{SOA Match} and \textbf{SIA Match} under different FD and semantic input.
It can be seen that SOA matches are more accurate and robust against semantic precision compared to SIA.} This can be attributed to the better stability of the centered object semantic against various matching noises. 
{On the other hand, the precision of SAM is limited by semantic precision. As shown in the table, the accuracy of area matching decreases with semantic accuracy, and the decrease is more significant at large FD. This limitation is discussed in detail in Sec.~\ref{sec:al}.}
In addition, both two areas are frequently involved in matching and sufficient number of area matches is also important for downstream tasks, indicating their importance for SGAM.

\begin{table}[t] 
\caption{\textbf{Area matching performance of two SAM areas and GP.} We construct area matching experiments on ScanNet for matching of two semantic areas, along with GP integrated with four point matchers. {The effect of two different semantic inputs is also evaluated.} AOR and AMP (with threshold $t=0.7$) under different matching difficulties (each with 1500 image pairs) are reported along with the area number per image {(Num)}. {The \textbf{best} and \uline{second} results under each semantic input setting and FD setting are highlighted.}}
\label{tab_SAM_GP_ab}
\resizebox{\linewidth}{!}{
\begin{tabular}{cllcllcllcl}
\toprule
& \multirow{2}{*}{Method}    & \multicolumn{3}{c}{FD@5} & \multicolumn{3}{c}{FD@10} & \multicolumn{3}{c}{FD@30} \\ 
    \cmidrule(r){3-5} \cmidrule(r){6-8} \cmidrule(r){9-11} 
  &  & AOR$\uparrow$      & AMP$\uparrow$    & Num   & AOR$\uparrow$      & AMP$\uparrow$    & Num   & AOR$\uparrow$      & AMP$\uparrow$    & Num    \\ \cmidrule(r){1-2} \cmidrule(r){3-5} \cmidrule(r){6-8} \cmidrule(r){9-11}
\multirow{6}*{\rotatebox{90}{GT Sem.}} & {SOA Match} & {85.94}    & 94.10      & 3.13      & 85.26   & 91.76      & 2.91       & 70.84    & 68.36      & 2.30       \\
 &  {SIA Match} & 83.67    & 91.91      & 2.38      & 83.50    & 84.35      & 2.01       & 66.94    & 62.17      & 1.26      \\ \cmidrule(r){2-2} \cmidrule(r){3-5} \cmidrule(r){6-8} \cmidrule(r){9-11}
 & GP\_ASpan & 86.59    & \uline{96.70}      & \multirow{4}{*}{0.26}      & 84.83    & \uline{89.59}      & \multirow{4}{*}{0.36}       & \uline{81.26}    & \uline{86.97}      & \multirow{4}{*}{0.50}      \\
 & GP\_QuadT & \textbf{87.86}    & \textbf{96.82}      &       & 84.98    & 88.47      &        & \textbf{82.37}    & \textbf{87.91}      &       \\
 & GP\_LoFTR & \uline{87.51}    & 95.73      &       & \textbf{87.42}    & \textbf{92.18}     &        & 73.81    & 86.48      & \\
 & GP\_COTR & 86.46    & 95.27      &      & \uline{86.58}    & 89.37      &      & 73.12    & 82.59      &        \\
 \midrule
 \multirow{6}*{\rotatebox{90}{{SEEM-L Sem.}}} & {SOA Match} & {\uline{86.33}}    & {\uline{89.94}}      & {3.35}      & {81.14}   & {81.22}      & {4.94}       & {72.25}    & {62.74}      & {2.62}       \\
 & {SIA Match} & {83.39}    & {83.46}      & {2.5}1      & {77.19}    & {72.53}      & {2.21}       & {65.85}    & {51.01}      & {1.76}      \\ \cmidrule(r){2-2} \cmidrule(r){3-5} \cmidrule(r){6-8} \cmidrule(r){9-11}
 & {GP\_ASpan} & {84.90}    & {\textbf{90.66}}      & \multirow{4}{*}{{0.57}}      & {\textbf{83.34}}    & {\textbf{84.51}}      & \multirow{4}{*}{{0.64}}       & {\textbf{75.43}}    & {63.02}      & \multirow{4}{*}{{1.61}}      \\
 & {GP\_QuadT} & {{85.03}}    & {{87.26}}      &       & {81.54}    & {82.06}      &        & {74.03}    & {63.16}      &       \\
 & {GP\_LoFTR} & {\textbf{87.23}}    & {\uline{89.94}}      &       & {\uline{82.59}}    & {\uline{83.91}}     &        & {\uline{74.25}}    & {\textbf{64.44}}      & \\
 & {GP\_COTR} & {{85.39}}    & {{89.65}}      &      & {80.73}    & {81.54}      &      & {73.87}    & {\uline{63.48}}      &        \\ \midrule
\end{tabular}
}
\vspace{-1.2em}
\end{table}


\subsection{Understanding GAM}

\textit{1) GP Precision.}
This section focuses on examining the area matching performance of GP on ScanNet~\cite{scannet} across three difficulty levels {and two semantic input (GT and SEEM-L)}, each consisting of 1500 image pairs.
The results are shown in Tab.~\ref{tab_SAM_GP_ab}.
{It can be seen that the area matching precision of GP surpasses that of SAM under different matching difficulties. 
Similar to SAM, the precision of GP decreases with inaccurate semantic input, but it can establish more accurate area matches than SAM in all semantic input cases.}
These observations confirm the effectiveness of GP.
Furthermore, the choice of point matchers influences the performance of GP, with improved point matching leading to increased accuracy in area matching.  
Notably, the doubtful area count per image indicates non-trivial semantic ambiguity within SAM, which becomes more prevalent with increasing matching difficulty.
In conclusion, GP plays a crucial role in SGAM by addressing semantic ambiguity and enhancing area matching performance.
The visualization of GP are shown in Fig. \ref{fig_QRGP} of the appendix.

\textit{2) Ablation Study of GR Parameter.}\label{sec_abl_phi}
In order to thoroughly examine the influence of the parameter $\phi$ in GR on pose estimation performance, we conduct experiments on ScanNet. We evaluate two difficulty levels, FD@10 and FD@30, each consisting of 1500 image pairs.
The obtained results are depicted in Fig. \ref{fig_phis}.
It is evident that the choice of $\phi$ has minimal impact when the frame difference is 10.
In scenes where the matching difficulty is not too high, an adequate number of area matches are identified. With a smaller $\phi$ value, accurate area matches are selected, (thanks to our GMC module which ensures the even distribution of matches) leading to improved performance.
However, as the difficulty of area matching increases under FD@30, a smaller $\phi$ value can cause point matches to be spatially concentrated, resulting in planar degeneration in certain cases. Hence, selecting an appropriate $\phi$ is crucial in challenging matching scenarios. Based on empirical findings, we establish $\phi=0.5$ as the default value.
We visualize the GR in Fig. \ref{fig_QRGR} of the appendix.

\begin{figure}[!t]
\centering
\includegraphics[width=\linewidth]{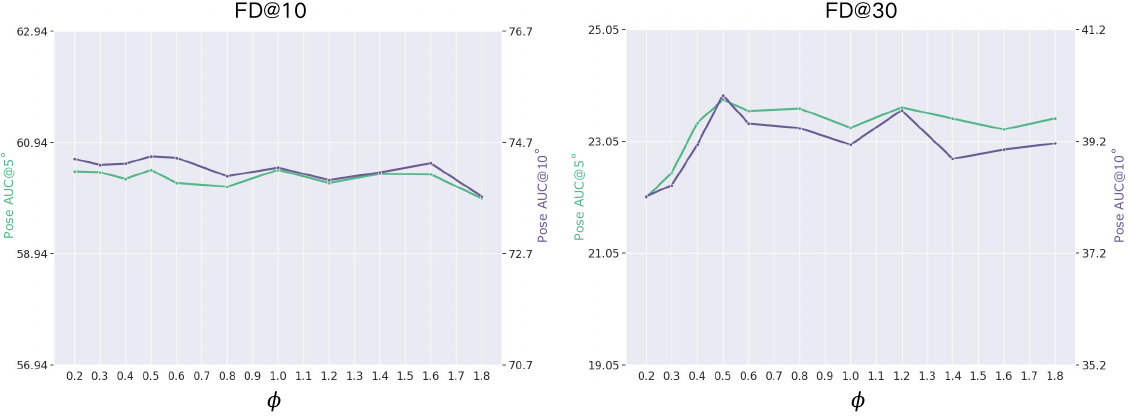}
\caption{\textbf{Ablation study of GR parameter $\phi$.} We report the pose estimation performance (AUC@$5^\circ/10^\circ$) of SGAM\_ASpan with various $\phi$ settings on ScanNet dataset under FD@$10$ (left) and FD@$30$ (right). Although smaller $phi$ brings more accurate area matches, it also aggregates point matches together which may result in planar degradation. Setting $\phi$ appropriately, therefore, is important especially in difficult matching scenes.}
\vspace{-1.2em}
\label{fig_phis}
\end{figure}

\subsection{Understanding Global Match Collection}\label{sec:GMC_ab}
{Global Match Collection (GMC) is another important module of our method, which ensures widely distributed point matches, particularly in less semantic scenes.}
To demonstrate the contribution of this module in our method, we conducted an ablation study on the ScanNet1500 benchmark.
The results are summarised in Fig. \ref{fig_gmc}.
The main parameter of our study is the size proportion threshold ($T_{SP}$), which represents the proportion of the image occupied by matched areas in the image pair. When $T_{SP}=0$, no GMC is performed, and when $T_{SP}=1$, all image pairs adopt the GMC module.
First, we demonstrate the relationship between $T_{SP}$ and the collection proportion, which represents the percentage of cases that adopt the GMC among all the pairs (Fig. \ref{fig_gmc}(a)). The collection proportion increases fast when size proportion threshold is small, indicating few-area-match pairs benefit from GMC.
Next, we display the pose estimation performance (AUC@$5^\circ/10^\circ$) under different size proportion thresholds (Fig. \ref{fig_gmc}(b)).
In the figure, the performance increases significantly as the threshold changes from $0.1$ to $0.3$. This is because the GMC module can significantly improve the distribution of matches, especially when there are only a few area matches established.
It is worth noting that even without GMC, our method can slightly improve the precision of the baseline (AUC@$5^\circ$: 25.89 vs. 25.78). However, the GMC module brings better performance. 
As the threshold further increases, the performance improvement levels off because the area matches already cover most of the overlap in the image pair.
Therefore, to achieve better performance while considering the computation cost of the GMC module (equal to one area match with images resized to the default size), the size proportion threshold can be set to 0.6.
Moreover, we visualize the affects of GMC in Fig. \ref{fig_gmc} (c),  where the distribution of matches is more uniform in the overlap area after applying GMC.

\begin{figure}[!t]
\centering
\includegraphics[width=\linewidth]{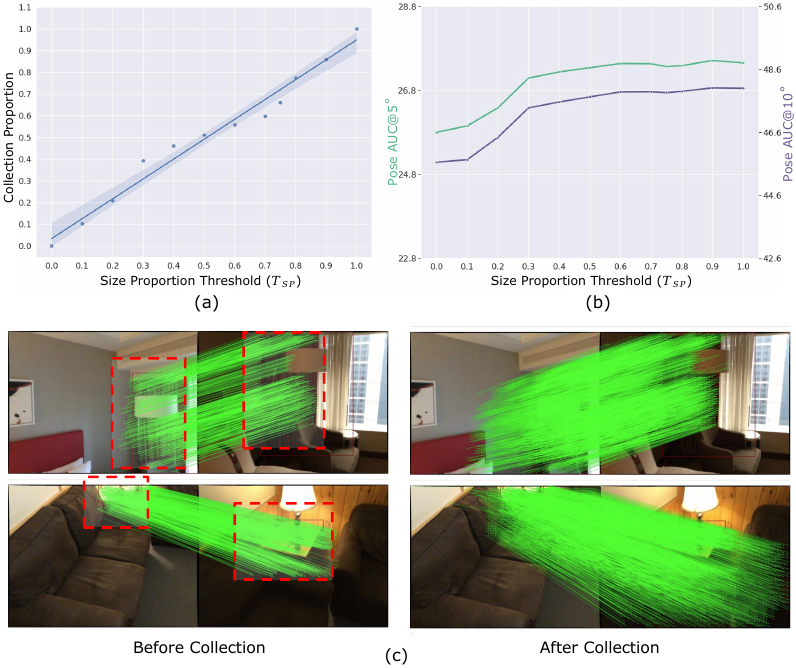}
\caption{\textbf{Ablation study of global match collection.} We conduct the ablation study to evaluate the effectiveness of global match collection module on the ScanNet1500 benchmark. \textbf{(a)} The relationship between the size proportion threshold and the number of cases using the global match collection module (collection proportion). \textbf{(b)} The relationship between the size proportion threshold and the pose estimation precision (Pose AUC@5$^\circ$/10$^\circ$). \textbf{(c)} The qualitative cases before and after global match collection. The red dash boxes indicate the area matches.}
\label{fig_gmc}
\vspace{-1.2em}
\end{figure}

\subsection{Running Time Comparison}
The proposed A2PM framework inherently decomposes the original matching task into \textbf{multiple} simpler matching tasks, leading to an inevitable increase in the time cost of SGAM. To demonstrate this, we conducted experiments on the YFCC100M dataset to compare the specific time costs among our method, the original point matching method, and another two-stage matching method, OETR. These experiments were performed on an Intel Xeon Silver 4314 CPU and a GeForce RTX 4090 GPU, and the results are presented in Table~\ref{tab_time_simp}. It is evident from the results that SGAM amplifies the time cost across all baselines, given that point matching is executed \textbf{multiple times} within the A2PM framework. However, SGAM demonstrates a similar time cost to OETR, although OETR conducts point matching \textbf{only once}.
This can be attributed to the semantic-aware search space utilized by SGAM, which results in a lightweight hand-crafted approach, contrasting with the computationally intensive nature of the learning approach in OETR.
Exploring the potential for parallel execution of multiple point matching in A2PM, akin to PATS (multiple patch matching operations are accelerated by CUDA), has the promise of substantially reducing the overall computational time. This avenue is one that we intend to investigate in our future research.
The detailed analyze of theoretical computational complexity of each part of SGAM is left in Sec.~\ref{sec:compute} of the appendix, along with the corresponding time costs.

\begin{table}[t]
\caption{\textbf{Running time comparison (s).} We construct experiments on YFCC100M to compare the time consuming between original point matchers and SGAM intergrated with them, along with another two-stage matching method.}
\label{tab_time_simp}
\resizebox{\linewidth}{!}{
\begin{tabular}{lcccccc}
\toprule
             & SP+SG & ASpan & QT   & LoFTR & COTR & PATS  \\ \midrule
time of original & 0.11 & 0.36  & 0.37 & 0.34  & 7.64 & 0.94 \\
time w/ SGAM & 1.04 & 1.42  & 1.49 & 1.37  & 22.52 & - \\
time w/ OETR & 0.84 & 1.03  & 1.14 & 0.95  & 8.32 & -  \\ \bottomrule
\end{tabular}}
\vspace{-1.2em}
\end{table}

\section{Conclusion}
To better incorporate semantic robustness into the coarse-to-fine feature matching, this study proposes semantic area matches as an intermediate search space for precise feature matching. The search space represents areas within images that exhibit prominent semantic features. By utilizing this search space, redundant computations are reduced, and the following point matcher receives high-resolution input, thereby improving overall matching performance. Aligned with the search space, the A2PM framework is introduced to hierarchically divide feature matching into two phases: first, establishing semantic area matches across the images, and then finding point matches within these area pairs. To implement the A2PM framework, we further propose SGAM method, comprising SAM and GAM, which leverages both semantic information and geometric constraints within the images. SAM conducts putative area matching based on large language model-powered semantic perception, while GAM, in conjunction with a point matcher, achieves precise area and point matches by ensuring geometric consistency. Extensive experiments validate the effectiveness of our approach, enhancing performance across sparse, semi-dense, and dense matching methods in point matching and downstream pose estimation tasks.

{
\normalem
\bibliographystyle{IEEEtran}
\bibliography{IEEEabrv, sgambib}

\begin{thebibliography}{10}
\providecommand{\url}[1]{#1}
\csname url@samestyle\endcsname
\providecommand{\newblock}{\relax}
\providecommand{\bibinfo}[2]{#2}
\providecommand{\BIBentrySTDinterwordspacing}{\spaceskip=0pt\relax}
\providecommand{\BIBentryALTinterwordstretchfactor}{4}
\providecommand{\BIBentryALTinterwordspacing}{\spaceskip=\fontdimen2\font plus
\BIBentryALTinterwordstretchfactor\fontdimen3\font minus \fontdimen4\font\relax}
\providecommand{\BIBforeignlanguage}[2]{{%
\expandafter\ifx\csname l@#1\endcsname\relax
\typeout{** WARNING: IEEEtran.bst: No hyphenation pattern has been}%
\typeout{** loaded for the language `#1'. Using the pattern for}%
\typeout{** the default language instead.}%
\else
\language=\csname l@#1\endcsname
\fi
#2}}
\providecommand{\BIBdecl}{\relax}
\BIBdecl

\bibitem{SLAMSurvey}
C.~Cadena, L.~Carlone, H.~Carrillo, Y.~Latif, D.~Scaramuzza, J.~Neira, I.~Reid, and J.~J. Leonard, ``Past, present, and future of simultaneous localization and mapping: Toward the robust-perception age,'' \emph{IEEE Transactions on Robotics}, vol.~32, no.~6, pp. 1309--1332, 2016.

\bibitem{COLMAP}
J.~L. Schonberger and J.-M. Frahm, ``Structure-from-motion revisited,'' in \emph{Proceedings of the IEEE conference on computer vision and pattern recognition}, 2016, pp. 4104--4113.

\bibitem{glampoints}
P.~Truong, S.~Apostolopoulos, A.~Mosinska, S.~Stucky, C.~Ciller, and S.~D. Zanet, ``Glampoints: Greedily learned accurate match points,'' in \emph{Proceedings of the IEEE/CVF International Conference on Computer Vision}, 2019, pp. 10\,732--10\,741.

\bibitem{matchsurvey}
J.~Ma, X.~Jiang, A.~Fan, J.~Jiang, and J.~Yan, ``Image matching from handcrafted to deep features: A survey,'' \emph{International Journal of Computer Vision}, vol. 129, no.~1, pp. 23--79, 2021.

\bibitem{sift}
D.~G. Lowe, ``Distinctive image features from scale-invariant keypoints,'' \emph{International journal of computer vision}, vol.~60, no.~2, pp. 91--110, 2004.

\bibitem{alike}
X.~Zhao, X.~Wu, J.~Miao, W.~Chen, P.~C. Chen, and Z.~Li, ``Alike: Accurate and lightweight keypoint detection and descriptor extraction,'' \emph{IEEE Transactions on Multimedia}, 2022.

\bibitem{superpoint}
D.~DeTone, T.~Malisiewicz, and A.~Rabinovich, ``Superpoint: Self-supervised interest point detection and description,'' \emph{Proceedings of the IEEE conference on computer vision and pattern recognition workshops}, pp. 224--236, 2018.

\bibitem{aslfeat}
Z.~Luo, L.~Zhou, X.~Bai, H.~Chen, J.~Zhang, Y.~Yao, S.~Li, T.~Fang, and L.~Quan, ``Aslfeat: Learning local features of accurate shape and localization,'' \emph{Computer Vision and Pattern Recognition (CVPR)}, 2020.

\bibitem{loftr}
J.~Sun, Z.~Shen, Y.~Wang, H.~Bao, and X.~Zhou, ``{LoFTR}: Detector-free local feature matching with transformers,'' \emph{{CVPR}}, 2021.

\bibitem{ASpan}
H.~Chen, Z.~Luo, L.~Zhou, Y.~Tian, M.~Zhen, T.~Fang, D.~McKinnon, Y.~Tsin, and L.~Quan, ``Aspanformer: Detector-free image matching with adaptive span transformer,'' in \emph{Computer Vision--ECCV 2022: 17th European Conference, Tel Aviv, Israel, October 23--27, 2022, Proceedings, Part XXXII}.\hskip 1em plus 0.5em minus 0.4em\relax Springer, 2022, pp. 20--36.

\bibitem{NCN}
I.~Rocco, M.~Cimpoi, R.~Arandjelovi{\'c}, A.~Torii, T.~Pajdla, and J.~Sivic, ``Neighbourhood consensus networks,'' \emph{Advances in neural information processing systems}, vol.~31, 2018.

\bibitem{dkm}
J.~Edstedt, I.~Athanasiadis, M.~Wadenbäck, and M.~Felsberg, ``{DKM}: Dense kernelized feature matching for geometry estimation,'' \emph{CVPR}, 2023.

\bibitem{MKPC}
H.~Song, Y.~Kashiwaba, S.~Wu, and C.~Wang, ``Efficient and accurate co-visible region localization with matching key-points crop (mkpc): A two-stage pipeline for enhancing image matching performance,'' \emph{arXiv preprint arXiv:2303.13794}, 2023.

\bibitem{OETR}
Y.~Chen, D.~Huang, S.~Xu, J.~Liu, and Y.~Liu, ``Guide local feature matching by overlap estimation,'' in \emph{Proceedings of the AAAI Conference on Artificial Intelligence}, vol.~36, no.~1, 2022, pp. 365--373.

\bibitem{pats}
Y.~L. Junjie~Ni, H.~L. Zhaoyang~Huang, Z.~C. Hujun~Bao, and G.~Zhang, ``Pats: Patch area transportation with subdivision for local feature matching,'' in \emph{The IEEE/CVF Computer Vision and Pattern Recognition Conference (CVPR)}, 2023.

\bibitem{sfd2}
F.~Xue, I.~Budvytis, and R.~Cipolla, ``Sfd2: Semantic-guided feature detection and description,'' in \emph{CVPR}, 2023.

\bibitem{topicfm}
K.~T. Giang, S.~Song, and S.~Jo, ``Topicfm: Robust and interpretable topic-assisted feature matching,'' in \emph{Proceedings of the AAAI Conference on Artificial Intelligence}, vol.~37, no.~2, 2023, pp. 2447--2455.

\bibitem{SEEM}
X.~Zou, J.~Yang, H.~Zhang, F.~Li, L.~Li, J.~Gao, and Y.~J. Lee, ``Segment everything everywhere all at once,'' \emph{arXiv preprint arXiv:2304.06718}, 2023.

\bibitem{orb}
E.~Rublee, V.~Rabaud, K.~Konolige, and G.~Bradski, ``Orb: An efficient alternative to sift or surf,'' \emph{2011 International conference on computer vision}, pp. 2564--2571, 2011.

\bibitem{r2d2}
J.~Revaud, C.~De~Souza, M.~Humenberger, and P.~Weinzaepfel, ``R2d2: Reliable and repeatable detector and descriptor,'' \emph{Advances in neural information processing systems}, vol.~32, 2019.

\bibitem{disk}
M.~Tyszkiewicz, P.~Fua, and E.~Trulls, ``Disk: Learning local features with policy gradient,'' \emph{Advances in Neural Information Processing Systems}, vol.~33, pp. 14\,254--14\,265, 2020.

\bibitem{d2net}
M.~Dusmanu, I.~Rocco, T.~Pajdla, M.~Pollefeys, J.~Sivic, A.~Torii, and T.~Sattler, ``D2-net: A trainable cnn for joint description and detection of local features,'' \emph{Proceedings of the ieee/cvf conference on computer vision and pattern recognition}, pp. 8092--8101, 2019.

\bibitem{keyNet}
A.~Barroso-Laguna, E.~Riba, D.~Ponsa, and K.~Mikolajczyk, ``Key. net: Keypoint detection by handcrafted and learned cnn filters,'' \emph{Proceedings of the IEEE/CVF International Conference on Computer Vision}, pp. 5836--5844, 2019.

\bibitem{rotation}
G.~B{\"o}kman and F.~Kahl, ``A case for using rotation invariant features in state of the art feature matchers,'' \emph{Proceedings of the IEEE/CVF Conference on Computer Vision and Pattern Recognition}, pp. 5110--5119, 2022.

\bibitem{eqkd}
J.~Lee, B.~Kim, and M.~Cho, ``Self-supervised equivariant learning for oriented keypoint detection,'' \emph{Proceedings of the IEEE/CVF Conference on Computer Vision and Pattern Recognition (CVPR)}, June 2022.

\bibitem{ng2022ninjadesc}
T.~Ng, H.~J. Kim, V.~T. Lee, D.~DeTone, T.-Y. Yang, T.~Shen, E.~Ilg, V.~Balntas, K.~Mikolajczyk, and C.~Sweeney, ``Ninjadesc: Content-concealing visual descriptors via adversarial learning,'' \emph{Proceedings of the IEEE/CVF Conference on Computer Vision and Pattern Recognition}, pp. 12\,797--12\,807, 2022.

\bibitem{pump}
J.~Revaud, V.~Leroy, P.~Weinzaepfel, and B.~Chidlovskii, ``Pump: Pyramidal and uniqueness matching priors for unsupervised learning of local descriptors,'' \emph{Proceedings of the IEEE/CVF Conference on Computer Vision and Pattern Recognition}, pp. 3926--3936, 2022.

\bibitem{decoupling}
K.~Li, L.~Wang, L.~Liu, Q.~Ran, K.~Xu, and Y.~Guo, ``Decoupling makes weakly supervised local feature better,'' \emph{Proceedings of the IEEE/CVF Conference on Computer Vision and Pattern Recognition}, pp. 15\,838--15\,848, 2022.

\bibitem{superglue}
P.-E. Sarlin, D.~DeTone, T.~Malisiewicz, and A.~Rabinovich, ``Superglue: Learning feature matching with graph neural networks,'' \emph{Proceedings of the IEEE/CVF Conference on Computer Vision and Pattern Recognition (CVPR)}, June 2020.

\bibitem{T-Net}
Z.~Zhong, G.~Xiao, L.~Zheng, Y.~Lu, and J.~Ma, ``T-net: Effective permutation-equivariant network for two-view correspondence learning,'' in \emph{Proceedings of the IEEE/CVF International Conference on Computer Vision (ICCV)}, October 2021, pp. 1950--1959.

\bibitem{OANet}
J.~Zhang, D.~Sun, Z.~Luo, A.~Yao, L.~Zhou, T.~Shen, Y.~Chen, L.~Quan, and H.~Liao, ``Learning two-view correspondences and geometry using order-aware network,'' in \emph{Proceedings of the IEEE/CVF International Conference on Computer Vision (ICCV)}, October 2019.

\bibitem{OR_TIP}
Y.~Xia and J.~Ma, ``Locality-guided global-preserving optimization for robust feature matching,'' \emph{IEEE Transactions on Image Processing}, vol.~31, pp. 5093--5108, 2022.

\bibitem{Msa}
L.~Zheng, G.~Xiao, Z.~Shi, S.~Wang, and J.~Ma, ``Msa-net: Establishing reliable correspondences by multiscale attention network,'' \emph{IEEE Transactions on Image Processing}, vol.~31, pp. 4598--4608, 2022.

\bibitem{PGF}
X.~Liu, G.~Xiao, R.~Chen, and J.~Ma, ``Pgfnet: Preference-guided filtering network for two-view correspondence learning,'' \emph{IEEE Transactions on Image Processing}, vol.~32, pp. 1367--1378, 2023.

\bibitem{cotr}
W.~Jiang, E.~Trulls, J.~Hosang, A.~Tagliasacchi, and K.~M. Yi, ``Cotr: Correspondence transformer for matching across images,'' in \emph{Proceedings of the IEEE/CVF International Conference on Computer Vision}, 2021, pp. 6207--6217.

\bibitem{sparseNCN}
I.~Rocco, R.~Arandjelovi{\'c}, and J.~Sivic, ``Efficient neighbourhood consensus networks via submanifold sparse convolutions,'' \emph{European conference on computer vision}, pp. 605--621, 2020.

\bibitem{dual-resolution}
L.~Xinghui, H.~Kai, L.~Shuda, and V.~P. Adrian, ``Dual-resolution correspondence networks,'' \emph{NIPS}, pp. 17\,346--17\,357, 2020.

\bibitem{tf}
A.~Vaswani, N.~Shazeer, N.~Parmar, J.~Uszkoreit, L.~Jones, A.~N. Gomez, {\L}.~Kaiser, and I.~Polosukhin, ``Attention is all you need,'' \emph{Advances in neural information processing systems}, vol.~30, 2017.

\bibitem{QT}
S.~Tang, J.~Zhang, S.~Zhu, and P.~Tan, ``Quadtree attention for vision transformers,'' \emph{ICLR}, 2022.

\bibitem{MVG_book}
R.~Hartley and A.~Zisserman, ``Multiple view geometry in computer vision,'' \emph{Cambridge university press}, 2003.

\bibitem{brief}
M.~Calonder, V.~Lepetit, M.~Ozuysal, and P.~Fua, ``Brief: Computing a local binary descriptor very fast,'' \emph{IEEE Transactions on Pattern Analysis and Machine Intelligence}, vol.~34, no.~7, pp. 1281--1298, 2011.

\bibitem{scannet}
A.~Dai, A.~X. Chang, M.~Savva, M.~Halber, T.~Funkhouser, and M.~Nie{\ss}ner, ``Scannet: Richly-annotated 3d reconstructions of indoor scenes,'' \emph{Proc. Computer Vision and Pattern Recognition (CVPR), IEEE}, 2017.

\bibitem{Matterport3D}
A.~Chang, A.~Dai, T.~Funkhouser, M.~Halber, M.~Niessner, M.~Savva, S.~Song, A.~Zeng, and Y.~Zhang, ``Matterport3d: Learning from rgb-d data in indoor environments,'' \emph{International Conference on 3D Vision (3DV)}, 2017.

\bibitem{KITTI}
Y.~Liao, J.~Xie, and A.~Geiger, ``Kitti-360: A novel dataset and benchmarks for urban scene understanding in 2d and 3d,'' \emph{IEEE Transactions on Pattern Analysis and Machine Intelligence}, 2022.

\bibitem{yfcc100m}
B.~Thomee, D.~A. Shamma, G.~Friedland, B.~Elizalde, K.~Ni, D.~Poland, D.~Borth, and L.-J. Li, ``{YFCC100M}: The new data in multimedia research,'' \emph{Communications of the {ACM}}, vol.~59, no.~2, pp. 64--73, 2016.

\bibitem{focal}
J.~Yang, C.~Li, X.~Dai, and J.~Gao, ``Focal modulation networks,'' \emph{Advances in Neural Information Processing Systems (NeurIPS)}, 2022.

\bibitem{COCO}
T.-Y. Lin, M.~Maire, S.~Belongie, J.~Hays, P.~Perona, D.~Ramanan, P.~Doll{\'a}r, and C.~L. Zitnick, ``Microsoft coco: Common objects in context,'' in \emph{Computer Vision--ECCV 2014: 13th European Conference, Zurich, Switzerland, September 6-12, 2014, Proceedings, Part V 13}.\hskip 1em plus 0.5em minus 0.4em\relax Springer, 2014, pp. 740--755.

\end{thebibliography}
}

\clearpage

\section{Appendix}

\subsection{Implementation Details}\label{sec:imp_de}

{\textit{1) Parameter Setting.} 
In the SAM, two semantic object areas with center distance (Sec.~\ref{sec:soam}) less than $100$ pixels are fused. 
{The parameter is set for sparse detection results.}
The multiscale ratios are $[0.8, 1.2, 1.4]$ for the scale invariance enhancement of two area descriptors, which aim to achieve more semantic specificity.
The thresholds in $SOA$ and $SIA$ matching are set as $T_{H}=0.5$, $T_{l}=0.75$ and $T_{da}=0.2$, which are smaller for more restricted matching.
In SIA detection, the top layer reduce ratio in semantic pyramid is $r\!=\!8$, which is a trade-off between detection efficient and accuracy.
We empirically set the $\phi$ for $T_{GR}$ as $0.5$ for ScanNet and $1.0$ for other datasets, the ablation study for which can be seen in Sec.~\ref{sec_abl_phi}. The $T_{SP}$ is set as $0.6$ for ScanNet and $0.3$ for other datasets, based on ablation study in Sec.~\ref{sec:GMC_ab}.}

\textit{2) Area Size.} 
{The area size is also the input size for the point matcher embedded in SGAM. In practice, the default area size is set as $256\times256$ for ScanNet, $480\times 480$ for Matterport3D, YFCC, KITTI and $640\times 640$ for ScanNet1500. To achieve better performance in ScanNet1500, we fine-tuning the semi-dense and dense baselines in $640\times 640$ input on ScanNet following \cite{loftr,dkm}. For the matched semantic object areas (SOAs), the sizes are first expanded from the bounding boxes of objects to match the width-height ratio of the default size. Then the SOAs are cropped from the original images and resized to the default size before being entered into the point matcher. After matching, the correspondences outside the object bounding boxes are filtered out. For semantic intersection areas (SIAs), their size is related to the detection window, which is first set with the default size.
To further mitigate the scale issue between images, we first match SOAs, which are robust to scale variation as the actual sizes of objects are fixed. Then, we adjust the detection windows of SIA in two images using the average size variation of SOA bounding boxes to obtain areas with consistent scales.}

\begin{table}[t]
\centering
{\revised
\caption{{\textbf{Symbol Table.} The table provides a comprehensive list of symbols used in the paper, and brief descriptions for each.}}

\label{table_sym}
 \resizebox{\linewidth}{!}{
\begin{tabular}{ll}
\toprule
 \textbf{Symbol}                                 & \textbf{Description}                                                            \\ \midrule
$I_i$                             & Input image pair, $i\in \{0,1\}$                                                                                  \\ \hdashline \noalign{\vskip 0.5ex}
$I^s_i$                                & Semantic segmentation of $I_i$, $i\in \{0,1\}$.                                                                   \\ \hdashline \noalign{\vskip 0.5ex}
$\mathcal{M}_A$                        & A2PM framework. \\ \hdashline \noalign{\vskip 0.5ex}       
$AM$                                     & Area Matching method. \\ \hdashline \noalign{\vskip 0.5ex}        
$PM$                                     & Point Matching method. \\ \hdashline \noalign{\vskip 0.5ex}    
$\alpha_i$                             & An area in $I_0$ with $i$ as the index. \\ \hdashline \noalign{\vskip 0.5ex}    
$\beta_i$                              & An area in $I_1$ with $i$ as the index.  \\ \hdashline \noalign{\vskip 0.5ex}

$\pi(i)$                              & Index mapping between matched areas. \\ \hdashline \noalign{\vskip 0.5ex}
 
\multirow{2}{*}{$\pi_l(i)$}                             & $\pi(i)$ with index $l$, \\
& indicating the $l$-th area matching possibility.\\ \hdashline \noalign{\vskip 0.5ex}

$\mathcal{A}_{i,\pi(i)}$                         & $\mathcal{A}_{i,\pi(i)}=(\alpha_i, \beta_{\pi(i)})$, a matched area pair.                                    \\ \hdashline \noalign{\vskip 0.5ex}

$(p,q)$                                & A matched point pair, $q\in I_0,p \in I_1$.  \\ \hdashline \noalign{\vskip 0.5ex}

$\mathcal{P}_i$ & A set of $M$ point matches, \\
\multirow{2}{*}{$=\{(q_i^m,p_i^m)\}_m^M$} & $i$ is the index of point match set, \\
& $m$ is the index of point pair inside the set.               \\ \hdashline \noalign{\vskip 0.5ex}

$F_i$                                  & Fundamental matrix with index $i$. \\ \hdashline \noalign{\vskip 0.5ex}

$d_{i,j}=D(F_i,\mathcal{P}_j)$         & Sampson distance calculated by $F_i$ and $\mathcal{P}_j$. \\ \hdashline \noalign{\vskip 0.5ex}
\multirow{2}{*}{$\hat{d}^m_{i,i}$}                        & Single match Sampson distance, \\
& calculated by $F_i$ and a point pair $(q_i^m,p_i^m)$. \\ \hdashline \noalign{\vskip 0.5ex}
$G_{\mathcal{A}_{i,\pi(i)}}$                     & Geometry consistency of $\mathcal{A}_{i,\pi(i)}$. \\ \hdashline \noalign{\vskip 0.5ex} 

$SAM$                                  & Semantic Area Matching module.                                                                            \\ \hdashline \noalign{\vskip 0.5ex}

$SOA$                                  & Semantic Object Area.                                                                            \\ \hdashline \noalign{\vskip 0.5ex}

$T_{H}$                         &  Threshold for match rejection in $SOA$ matching.                                  \\ \hdashline \noalign{\vskip 0.5ex}

$T_{da}$                         &  Threshold for the doubtful area.                                  \\ \hdashline \noalign{\vskip 0.5ex}

$SIA$                                  & Semantic Intersection Area.                                                                            \\ \hdashline \noalign{\vskip 0.5ex}

$T_{l}$                         &  Threshold for match rejection in $SIA$ matching.                                  \\ \hdashline \noalign{\vskip 0.5ex}

\multirow{2}{*}{$GP_{PM}$}                             & Geometric area match Predictor, \\ 
&combined with point matcher $PM$.                                                  \\ \hdashline \noalign{\vskip 0.5ex}
\multirow{2}{*}{$GR_{PM}$}                              & Geometric area match Rejector, \\
&combined with point matcher $PM$. \\ \hdashline \noalign{\vskip 0.5ex}
$GMC_{PM}$                                  & Global Match Collection module with $PM$.   \\ \hdashline \noalign{\vskip 0.5ex}
$T_{GR}$                               & Threshold of geometry consistency in $GR$. \\ \hdashline \noalign{\vskip 0.5ex}
\multirow{2}{*}{$\mathcal{A}\mathcal{S}_{l}$}                               & $\mathcal{A}\mathcal{S}_l = \{\mathcal{A}_{i\pi_l(i)}\}_{i}^{R}$ , a set of $R$ area matches \\
& with index $l$ related to $\pi_l(i)$.  \\ \hdashline \noalign{\vskip 0.5ex}
$G_{As_{l}}$                           & Geometry consistency of area match set $\mathcal{A}\mathcal{S}_{l}$. \\ \hdashline \noalign{\vskip 0.5ex}
\multirow{2}{*}{$SP_{\{\mathcal{A}_{i,\pi(i)}\}_i}$}                               & Size Proportion of area matches $\{\mathcal{A}_{i,\pi(i)}\}_i$ \\
 & in the image pair.                                                               \\ \hdashline \noalign{\vskip 0.5ex}
$T_{SP}$                               & Threshold of size proportion in $GMC$.                                                        \\ \bottomrule
\end{tabular}
}
}
\end{table}

\textit{3) Semantic Noise Filtering.} 
Our method takes semantic segmentation images as input, obtained from SOTA semantic segmentation methods or ground truth labels.
However, even if provided with manual labels, these inputs can still contain semantic labeling errors.
Thus, the semantic noise filtering needs to be performed in SAM.
Specifically, in semantic object area detection, objects smaller than ${1}/{100}$ image size is ignored and the semantic surrounding descriptor neglects semantics with fewer than $20$ continuous pixels at the boundary.
In semantic intersection area detection, semantics smaller than ${1}/{64}$ window size are filtered. The semantic labels with size less than ${1}/{64}$ area size are filtered out in the construction of semantic proportion descriptor. 

\subsection{{Ablative Study of Components}}
{
We also conducted a dedicated experiment to systematically decompose the components of our approach, evaluating the area matching and pose estimation performance. The threshold of AMP is set as $t=0.7$. We sample 1500 image pair with frame difference is 15 (FD@15) from ScanNet for this experiment. Our SGAM is combined with ASpan. The results are reported in Tab.~\ref{tab_dc_ab}. It can be seen that both area matching and pose estimation accuracy improve as the completeness of our method increases. This finding confirms the effectiveness of all the components within our SGAM approach.
}

\subsection{Ablation Study of Maximum Correspondence Number}
{Another concern about GMS is that it achieves more but may be duplicate correspondences, due to the overlaps between area matches and GMS.
However, as we uniformly sample no more than the \textit{Maximum Correspondence Number} (MCN) of points in the image space for pose estimation, duplicate correspondences are removed. Additionally, we conduct experiments on both ScanNet (FD@10, 1.5k image pairs) and YFCC100M, to investigate the influence of the MCN on the pose estimation, which is set as 500, 800 and 1000 respectively. The results are reported in Table~\ref{tab_pn_ab}. It is evident that the MCN have slight impact on pose estimation and SGAM demonstrates improvement across all settings.}

\begin{figure*}[!t]
\centering
\includegraphics[width=\linewidth]{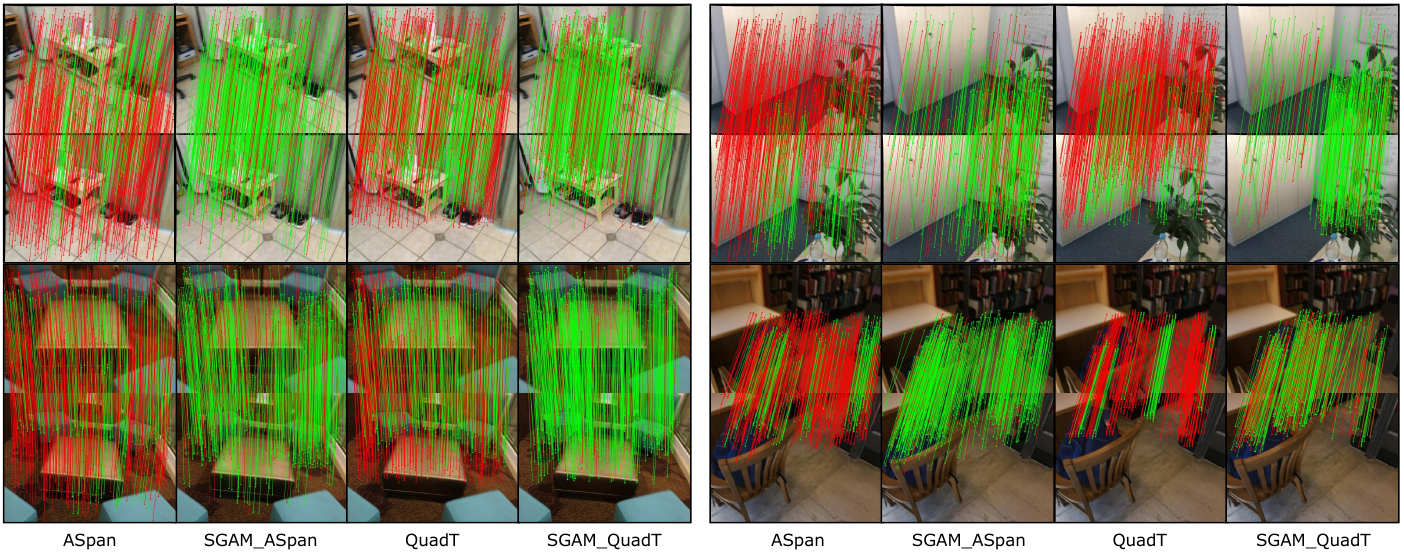}
\caption{\textbf{Qualitative Comparison on ScanNet.} The visual comparison between our method and two SOTA baselines.
The \textcolor{red}{wrong} and \textcolor{green}{correct} matches under the same threshold are labeled respectively.
The point matches we obtained possess much higher precision as well as uniform distribution. }
\vspace{-1.0em}
\label{fig_qrsn}
\end{figure*}

\begin{table}[t]
\centering
{\revised
\caption{{\textbf{Ablation study of Components.} We conduct the decomposing component experiment of \colorbox{colorFst}{SGAM\_ASpan} on ScanNet with FD@15, using 1k image pairs. The area matching and pose estimation performance are reported. The numbers of area matches (Num) are also reported. } }
\label{tab_dc_ab}
\resizebox{\linewidth}{!}{
\begin{tabular}{ccccccccccc}
\toprule
SOA & SIA & GP & GR & GMC & AOR~$\uparrow$   & AMP~$\uparrow$ & Num & AUC@5~$^\circ$$\uparrow$ & AUC@10~$^\circ$$\uparrow$ & AUC@20~$^\circ$$\uparrow$ \\ \cmidrule(r){1-5} \cmidrule(r){6-11}
\CheckmarkBold &     &    &    &     & 79.39 & 78.88   & 2.74    & 32.96 & 43.77  & 53.80   \\
    & \CheckmarkBold  &    &    &     & 74.64 & 69.42   & 2.11    & 31.72 & 40.66  & 46.89  \\
\CheckmarkBold  & \CheckmarkBold  &    &    &     & 77.41 & 74.95   & 4.53    & 34.02 & 48.26  & 52.94  \\
\CheckmarkBold  & \CheckmarkBold  & \CheckmarkBold &    &     & 79.03 & 76.30    & 5.57    & 39.54 & 52.97  & 64.54  \\
\CheckmarkBold  & \CheckmarkBold  &    & \CheckmarkBold &     & 79.01 & 78.00      & 3.73    & 49.24 & 62.01  & 73.32  \\
\CheckmarkBold  & \CheckmarkBold  & \CheckmarkBold & \CheckmarkBold &     & 79.18 & 78.27   & 4.01    & 49.73 & 62.58  & 73.17  \\
\rfs
\CheckmarkBold  & \CheckmarkBold  & \CheckmarkBold &  \CheckmarkBold & \CheckmarkBold  & 79.18 & 78.27   & 4.01    & 50.50  & 63.38  & 74.75  \\ \bottomrule
\end{tabular}
}
}
\end{table}

\begin{table}[t]
\centering
{\revised
\caption{{\textbf{Comparison experiment on maximum correspondence number (MCN).} The pose estimation results on ScanNet with FD@10 and YFCC100M are reported. \colorbox{colorFst}{Our method} is combined with ASpan and takes semantic input from SEEM-T. We show the improvement of our method on the accuracy of ASpan.}}
\label{tab_pn_ab}
\resizebox{\linewidth}{!}{
\begin{tabular}{ccllllll}
\toprule
\multirow{2}{*}{MCN} & \multirow{2}{*}{Method} & \multicolumn{3}{c}{ScanNet-FD@10} & \multicolumn{3}{c}{YFCC100M} \\ \cmidrule(l){3-5} \cmidrule(l){6-8}
                          &                         & AUC@5$^\circ$$\uparrow$    & AUC@10$^\circ$$\uparrow$    & AUC@20$^\circ$$\uparrow$    & AUC@5$^\circ$$\uparrow$   & AUC@10$^\circ$$\uparrow$   & AUC@20$^\circ$$\uparrow$  \\ \midrule
\multirow{2}{*}{500}      & ASpan                   & 58.51    & 70.42     & 79.84     & 38.96   & 59.35    & 75.54   \\
                           & \fs SGAM\_ASpan             & \fs 60.78$_{+3.88\%}$    & \fs 74.24$_{+5.42\%}$     & \fs84.53$_{+5.87\%}$     & \fs39.77$_{+2.08\%}$   & \fs60.24$_{+1.50\%}$    & \fs76.21$_{+0.89\%}$   \\ \midrule
\multirow{2}{*}{800}      & ASpan                   & 58.02    & 68.45     & 78.41     & 39.35   & 59.72    & 75.67   \\
                           & \fs SGAM\_ASpan             & \fs 60.56$_{+4.38\%}$    & \fs 71.17$_{+3.97\%}$     & \fs 79.84$_{+1.82\%}$     & \fs 40.05$_{+1.78\%}$   & \fs 60.53$_{+1.36\%}$    & \fs 76.54$_{+1.15\%}$   \\ \midrule
\multirow{2}{*}{1000}     & ASpan                   & 57.81    & 67.66     & 77.59     & 39.18   & 59.54    & 75.60   \\
                          & \fs SGAM\_ASpan             & \fs 60.24$_{+4.20\%}$    & \fs 72.65$_{+7.38\%}$     & \fs 80.77$_{+4.10\%}$     & \fs 40.55$_{+3.50\%}$   & \fs 60.84$_{+2.18\%}$    & \fs 76.78$_{+1.56\%}$  \\ \bottomrule
\end{tabular}
}
}
\vspace{-1.0em}
\end{table}

\begin{figure}[!t]
\centering
\includegraphics[width=\linewidth]{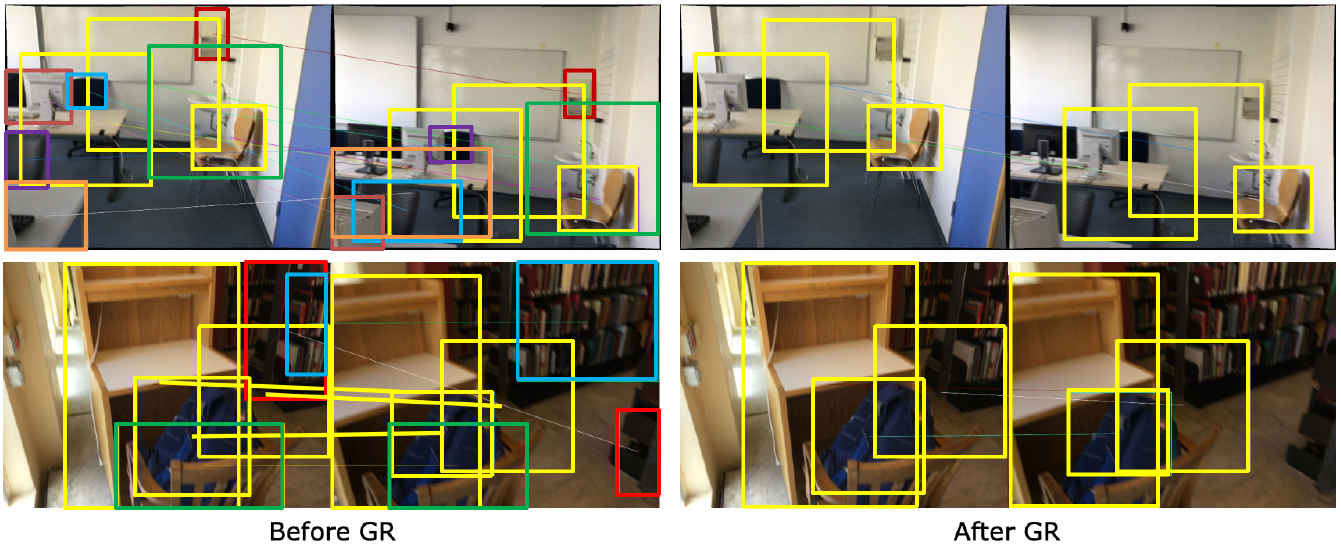}
\caption{\textbf{Visualization of GR.} {After GR, many false and inaccurate area matches (boxes of the same color) are rejected. Only reliable area matches (highlighted by {\color{yellow}yellow} boxes) are left, leading to high matching accuracy for point matching.}}
\vspace{-1.2em}
\label{fig_QRGR}
\end{figure}

\begin{figure}[!t]
\centering
\includegraphics[width=\linewidth]{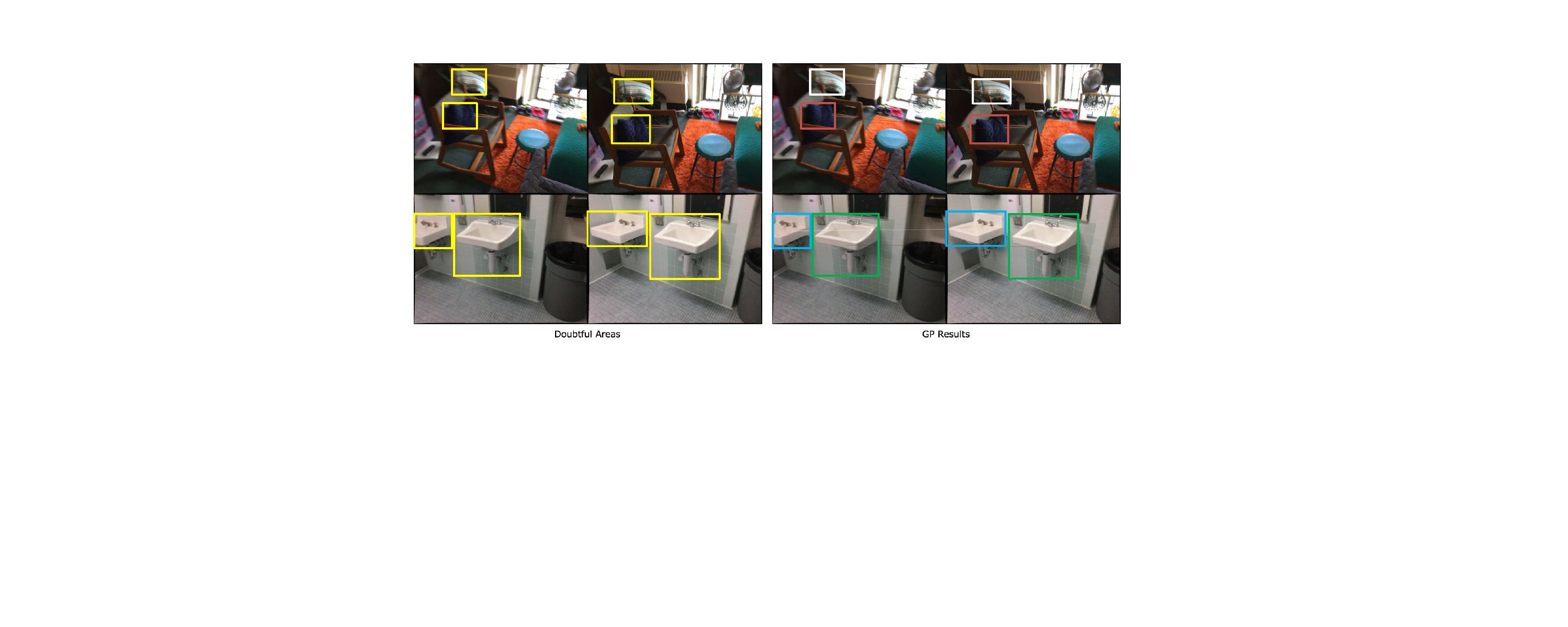}
\caption{\textbf{Visualization of GP.} {The cases processed by GP, which can predict the true matches (boxes of the same color) from the doubtful candidates ({\color{yellow}yellow} boxes) in semantic ambiguity.}}
\vspace{-1.2em}
\label{fig_QRGP}
\end{figure}

\begin{figure*}[!t]
\centering
\vspace{-0.5in}
\includegraphics[width=\linewidth]{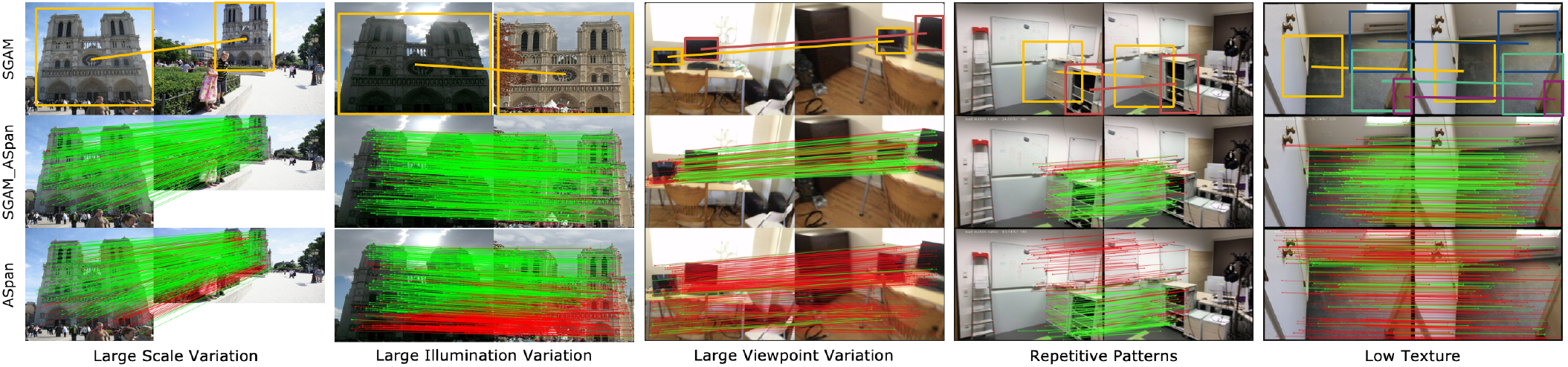}
\caption{{\textbf{Qualitative Comparison on challenging scenes.} The visual comparison between our method and ASpan in challenging scenes.
The \textcolor{red}{wrong} and \textcolor{green}{correct} matches under the same threshold are labeled respectively.
}}  
\vspace{-1.2em}
\label{fig_qr_expc}
\end{figure*}

\subsection{{Discussion on the Computational Complexity}}\label{sec:compute}

In this section, we discuss in detail the theoretical computational complexity of each component of our proposed approach, including SAM, GP and GR.
Moreover, we conduct an experiment to count the average running time per image pair of each component of our method in practice. We collect 1500 sets of image pairs from ScanNet with FD@10 for this experiment. Four baseline point matchers are combined for time comparison. The results are reported in Tab.~\ref{tab_time}. This experiment is run on a Intel Xeon Silver 4314 CPU and a GeForce RTX 4090 GPU.

\begin{table}[t]
\centering
\resizebox{\linewidth}{!}{
\begin{threeparttable}
\caption{{\textbf{Time Consumption Comparison.} The experiment is conducted on ScanNet with FD@10. The time consumption of each component of our method with specific input size is reported. Different time consumption comes from different baselines coupled with our method are investigated as well. The time of baselines are also reported.}}
\label{tab_time}
\begin{tabular}{ccccccc}
\toprule
\multirow{2}{*}{Time/s} & \multicolumn{6}{c}{Input Size}                                          \\ \cmidrule(l){2-7}
                      & 640$\times$480  & \multicolumn{4}{c}{256$\times$256}  & 640$\times$480 \\ \cmidrule(r){1-1} \cmidrule(r){2-2} \cmidrule(r){3-6} \cmidrule(r){7-7} 
PMer\tnote{1}                    & SAM                    & GP    & GR    & GMC   & SGAM    & PM\tnote{2}       \\ \cmidrule(r){1-1} \cmidrule(r){2-2} \cmidrule(r){3-6} \cmidrule(r){7-7} 
ASpan                   & \multirow{4}{*}{0.62} & 0.042 & 0.20  & 0.021 & 0.88   & 0.19    \\
QuadT                      &                        & 0.040 & 0.17  & 0.023 & 0.85   & 0.18    \\
LoFTR                   &                        & 0.041 & 0.19  & 0.018 & 0.86   & 0.18    \\
COTR                    &                        & 2.54  & 23.85 & 2.13  & 29.14  & 56.04  \\ \bottomrule
\end{tabular}
\begin{tablenotes}
\footnotesize
\item[1] point matcher incorporated by SGAM;
\item[2] point matching on the entire images;
\end{tablenotes}
\end{threeparttable}
}
\end{table}

\textit{1) Computational Complexity of SAM.}
SAM mainly includes area detection and description for two semantic areas (SOA and SIA). For SOA, the size of the algorithm is related to the number of semantic categories in the image pair ($N_{sem}$). Thus the computational complexity for this part is $O(N_{sem})$. The SIA part involves a sliding window algorithm on image. Its computational complexity is $O((W_I - W_w)*(H_I - H_w)/s^2)$, where $W_I,H_I$ are the \textit{Width} and \textit{Height} of the \textit{Image}, $W_w,H_w$ are \textit{Width} and \textit{Height} of \textit{Window} and $s$ is the sliding step. As the window size is large (please refer to Sec.~\ref{sec:siam}), the time consumption of this part is acceptable. It can be seen in Tab.~\ref{tab_time} that SAM takes $0.62$s to perform area matching in a pair of $640\times 480$ images. The speed can be further enhanced, as the current code has not yet undergone optimization for speed.

\textit{2) Computational Complexity of GP.}
Given $H$ doubtful areas in $I_0$ and $R$ in $I_1$, the GP can determine area matches through point matching within areas.
Relying on correspondences, however, leads to multiple times point matching for all area match possibilities ($H \times R$ times) in the single image pair. At the same time, the calculation of $P({As}_l)$ in Eq.~\ref{eq:GP} also need to be performed $L = \frac{H!}{(H-R)!}$ times. Thus, for $P({As_l})$ calculation, its computational complexity varies from $O(N)$ (when $R = 1$) to $O(N!)$ (when $R = H - 1$), depending on the area number. However, as we only perform this prediction when semantic ambiguity occurs, the practical time cost is acceptable. This can be seen in Tab.~\ref{tab_time}. The speed of GP is determined by the point matcher used, with its time consumption being comparable to that of single-image matching (refer to the GMC column, representing the time cost of single image matching).

\textit{3) Computational Complexity of GR.}
In GR, point matching inside area matches are performed to compute geometry consistency. This inside-area matching is the key of our A2PM framework, which is equivalent to decomposing a matching problem (the full-image matching) into multiple matching problems (the inside-area matching). Thus the time consumption inevitably rises, when the input resolution of SGAM and original point matching is the same, as shown in Tab.~\ref{tab_time} (SGAM column vs. GMC column). 
The computational complexity of widely-used vanilla Transformer in SOTA matching method~\cite{cotr} is $O(N^2)$, where $N$ is the input size. Thus, this decomposing of matching in A2PM is more efficient than direct matching, when the area size is smaller enough than the original image size. 
Specifically, take point matching using Transformer~\cite{cotr} as an example, whose computational complexity is $O(N^2)$ and $N$ is the input size.
Suppose the image size is $W_I\times H_I$, area size is $W_a\times H_a$ and area match number is $N_a$. Then the computational complexity of original point matcher is $O((W_I\times H_I)^2)$, while the computational complexity of A2PM is $O(N_a\times(W_a\times H_a)^2)$. 
Thus, when the matching area size and image size are the same, the time cost rises with the area number.
our A2PM is more effective than the original point matcher, when $(W_I\times H_I)^2/ (W_a\times H_a)^2 \geq N_a$. The results in Table \ref{tab_time} substantiate this claim. When the COTR is employed, SGAM\_COTR with a $256\times 256$ input ($29.14$s) demonstrates higher speed than the original COTR with a $640\times 480$ input ($56.04$s), attributable to its second-order computational complexity. Some recent methods use linear Transformer whose computational complexity is $O(N)$. In this case, our A2PM is more effective than the original point matcher, when $(W_I\times H_I)/ (W_a\times H_a) \geq N_a$.

\subsection{{Advantages and Limitations}}\label{sec:al}
{As semantic possesses consistency against various matching noises, e.g. illumination, viewpoint and scale changes between images, our SGAM is able to find accurate area matches in challenging scenes. Then, the precision of inside-area point matching is significantly boosted, due to noise removal and higher resolution of these areas, which is the main advantage of our method. We also provide qualitative comparison results for five hard scenarios in Fig.~\ref{fig_qr_expc} to demonstrate the superiority of SGAM.
However, heavy reliance on semantics also results in limitations of SGAM.
First, the semantic segmentation accuracy impacts the performance of our method. Especially for area matching, our SAM and GP exhibits non-negligible decreases in precision (Tab.~\ref{table_amp} and Tab.~\ref{tab_SAM_GP_ab}). This is because the area detection and description in SAM both assume accurate semantic input. Thus the under-splitting, over-splitting, and multiple semantics in inferior semantic segmentation leads to reduced performance of SAM and large doubtful area numbers for GP. 
However, it is noteworthy that SGAM is still able to improve the point matching and pose estimation performance with SEEM-L/T in our experiments. This highlights the fact that the advantages of SGAM are still significant in the presence of less accurate semantic inputs and implies the potential of our A2PM framework.
Meanwhile, scenarios involving mirroring may also lead to area mismatches, which may further introduce incorrect point matches. This situation also reveals the importance of GAM; as long as most of the area matches are correct, GAM can screen out the false matches, which greatly reduces the impact of these specific scenarios on the performance of our methods.

The second limitation of SGAM is related to the spatial granularity of semantic categories.}
For example, when a single semantic entity dominates the image, it is difficult for SAM to find areas with clustered features.
Hence the effectiveness of SGAM is restricted in some scenes, such as YFCC100M.
However, in such scenes, the A2PM framework still benefits feature matching, but the area matches need to be established by other approaches, like overlap estimation~\cite{OETR,MKPC}.
In our future work, we will focus on area matching without semantic prior, which may work well in more general application scenes.

\end{document}